\documentclass[lettersize,journal]{IEEEtran}
\usepackage{amsmath,amsfonts}
\usepackage{array}
\usepackage[caption=false,font=normalsize,labelfont=sf,textfont=sf]{subfig}
\usepackage{textcomp}
\usepackage{stfloats}
\usepackage{url}
\usepackage{verbatim}
\usepackage{graphicx}
\usepackage{cite}
\usepackage[numbers]{natbib}
\usepackage{bibunits}
\defaultbibliographystyle{IEEEtranN}
\defaultbibliography{diffusion-robust}
\usepackage[utf8]{inputenc}
\usepackage[T1]{fontenc}
\usepackage{hyperref}
\usepackage{url}
\usepackage{booktabs}
\usepackage{amsfonts}
\usepackage{nicefrac}
\usepackage{microtype}
\usepackage{xcolor}
\usepackage{amsmath}
\usepackage{amsthm}
\usepackage{amssymb}
\usepackage{bm}
\usepackage{graphicx}
\usepackage{multirow}
\usepackage{multicol}
\usepackage{enumerate}
\usepackage{paralist}
\usepackage{tablefootnote}
\newtheorem{definition}{Definition}

\newtheorem{model}{Model}
\newtheorem{assumption}{Assumption}
\DeclareMathOperator{\tr}{tr}
\usepackage{booktabs}
\usepackage{xcolor}
\usepackage{dsfont}
\usepackage{wrapfig}
\usepackage{pifont}
\newcommand{\cmark}{\ding{51}}
\newcommand{\xmark}{\ding{55}}
\usepackage{mathtools}
\usepackage{algorithm}
\usepackage{algpseudocode}

\newcommand{\parallelsum}{\mathbin{\!/\mkern-5mu/\!}}

\begin{document}
	
	\title{Demystifying Adversarial Robustness in Diffusion Models: Compression, Randomness, and Geometry}
	
	\author{Liu~Yuezhang~and~Xue-Xin~Wei
		\thanks{Liu Yuezhang and Xue-Xin Wei are with The University of Texas
			at Austin, Austin, TX 78712, USA (e-mail: lyz@utexas.edu;
			weixx@utexas.edu).}
		\thanks{This research was supported by a Sloan Research Fellowship
			(to X.-X. Wei) and NSF 2318065.}}

\maketitle

\begin{abstract}
	Recent studies suggest that diffusion models significantly improve the empirical adversarial robustness of deep neural network models. While intuitive explanations have been proposed, the mechanisms underlying diffusion-based robustness remain largely unclear. This work aims to demystify how diffusion models improve adversarial robustness. We observe that diffusion models surprisingly increase the $\ell_p$ distance to clean samples, thus rejecting the hypothesis that purification denoises perturbed images closer to the clean ones. Next, we provide a unifying account of the robustness improvement in diffusion-based purification by decomposing it into two sources: (i) gradient masking induced by randomness; (ii) compression of the image space. First, we find that the purified images are heavily influenced by the internal randomness of diffusion models. This randomness leads to gradient masking that cannot be removed by the previously proposed remedy, i.e., expectation-over-transformation (EOT). The improvement in robustness due to randomness is determined by the cosine similarity of the optimal vs. empirical attack directions, as predicted by a hyperspherical cap model of the adversarial regions.  
Second, we find that, when fixing the randomness, diffusion models substantially compress the image space. Importantly, we discover a lawful relationship between the model’s ability to compress the image space and the genuine adversarial robustness gain. Further theoretical analyses show that convergent score fields encoded in diffusion models explain these compression effects.
Our findings reveal new insights into the mechanisms underlying diffusion-based purification, and offer guidance for developing more effective and principled adversarial purification systems.

\end{abstract}


\makeatletter
\global\let\oldaddcontentsline\addcontentsline
\global\renewcommand{\addcontentsline}[3]{}
\makeatother

\begin{bibunit}

\section{Introduction}
\IEEEPARstart{N}{eural} networks are vulnerable to small adversarial perturbations~\citep{szegedy2013intriguing, Goodfellow2014explaining}. The lack of robustness presents a fundamental problem of artificial learning (AI) systems.
Adversarial training~\citep{madry2017towards} has been proposed as an effective method to overcome this problem under certain scenarios~\citep{Shafahi2019adversarial,pang2020bag,Wang2021convergence}. However, research has found that training with a specific attack usually sacrifices the robustness against other types of perturbations~\citep{schott2018towards,ford2019adversarial,yin2019fourier}, indicating that adversarial training may overfit the attack rather than achieving an overall robustness improvement.

Adversarial purification represents an alternative promising path toward adversarial robustness. This approach typically uses generative models to purify the image before passing it to a classifier~\citep{song2018pixeldefend,samangouei2018defense,Shi2021online,yoon2021adversarial,Ouyang2023improving}. The basic idea is to leverage the image priors learned in generative models to project adversarial perturbations back toward the image manifold. Intuitively, the performance of such purification procedures should depend on how well the generative models encode the probability distribution over natural images. Recently, adversarial purification based on diffusion models~\citep{ho2020denoising,song2020score} (DiffPure) was reported to show impressive improvements against various empirical attacks~\citep{nie2022diffusion}. The idea of using diffusion models as denoisers was further combined with the denoised smoothing framework~\citep{cohen2019certified, salman2020denoised} to improve certified robustness~\citep{Carlini2022, xiao2023dense}. However, some subsequent studies~\citep{lee2023robust,li2024adbm,Kassis2025diffbreak} reported that there was an overestimate of the robustness improvement from the DiffPure method. Meanwhile, there is a debate on the role that randomness in diffusion models plays in evaluating robustness~\cite{yuezhang2023dissecting,li2024adbm,Kassis2025diffbreak}. In sum, despite some promising empirical results, the precise mechanisms underlying the improvement in diffusion-based purification remain poorly understood~\cite{Chen2024opportunities}.
To address this basic gap, we systematically investigated how diffusion models improve adversarial robustness. Our work makes several contributions.

First, we demonstrate that the intuitive denoising hypothesis (\textit{i.e.}, diffusion models push the adversarially perturbed images towards clean samples)~\cite{nie2022diffusion} provided a poor explanation of the robustness improvement under diffusion models. In fact, diffusion models do the opposite -- they generally push the perturbed images further away from clean samples. In addition, the distributional alignment hypothesis (i.e., after purification, the distribution of
perturbed images becomes more closely aligned with the clean image distribution) also fails to explain the amount of robustness improvement across different diffusion models. 

Second, we find that empirical robustness obtained in diffusion-based purification comes from two distinct sources: (i) \textit{bona fide} robustness improvement, quantified by attacking diffusion models with fixed randomness; (ii) randomness-induced suboptimal gradients in generating the attacks. Crucially, we found that both terms are determined by interpretable factors. For (i), the genuine robustness improvement is predicted by how much diffusion models compress the image space, which is effectively characterized by the compression rate (a scalar variable). For (ii), the residual robustness due to randomness is determined by the cosine similarity between the actual attack and the optimal attack under fixed randomness, as predicted by a hyper-spherical cap model of the adversarial region. Together, for a given diffusion-based purification method, the empirical robustness is determined by these two factors. If one is primarily interested in the robustness of diffusion-based purification without randomness, it is primarily determined by the compression of the image space. 

Third, to understand why diffusion models compress the image space, we theoretically analyze the source of compression in the context of both simple Gaussian models and diffusion models. We find that compression originates from the convergent score field learned in diffusion models. The compression rate mentioned above captures off-manifold compression, while on-manifold perturbations remain largely preserved. 

Some of our results may be of interest in their own right beyond diffusion-based purification. In particular, the hyper-spherical cap model, which we proposed and empirically validated, is relevant for understanding the structure of adversarial regions of deep networks in general. It can be connected to prior work that studies the dimensionality of adversarial space~\cite{szegedy2013intriguing,tramer2017space}. Our analysis on the convergence of the score fields in diffusion models may be relevant for understanding the geometrical structure of natural image priors. 
This paper does not directly propose new procedures for increasing robustness-- our primary goal is to reveal mechanistically how diffusion-based purification works. However, we envision that an improved understanding of how diffusion-based purification improves robustness will provide guidance for the development of future approaches to increase the robustness of AI.

\section{Related Work and Preliminaries}
\label{sec:notation}
\paragraph{Generative models for adversarial purification} Unlike adversarial training which directly augments the classifier training with adversarial attacks, adversarial purification intends to first ``purify'' the perturbed image before classification. Generative models are usually utilized as the purification system, such as denoising autoencoder~\citep{Gu2014towards}, denoising U-Net~\citep{liao2018defense}, PixelCNN~\citep{song2018pixeldefend} and GAN~\citep{samangouei2018defense}.
Diffusion models~\citep{ho2020denoising,song2020score} achieve state-of-the-art performance on image generation, and represent a natural choice for adversarial purification. \citet{nie2022diffusion} proposed the DiffPure framework, which utilized both the forward and reverse process and achieved promising empirical robustness comparable to adversarial training on multiple benchmarks. Similar improvements were reported with guided diffusion models~\citep{wang2022guided}. These studies led to substantial interest in applying diffusion models for adversarial purification in various domains, including auditory data~\citep{wu2022defending} and 3D point clouds~\citep{sun2023critical}. Recently, other techniques, such as adversarial guidance~\citep{lin2024robust} and bridge models (ADBM)~\citep{li2024adbm}, were introduced to further enhance robustness. Beyond purification, diffusion models have also been used directly as robust generative classifiers~\citep{chen2024rdc}. To address the heavy sampling cost of multi-step purification, \citet{li2025adversarial} proposed a one-step guided diffusion model reported to maintain robustness while purifying in a single denoising step. Another line of research applies diffusion models to improve certified robustness~\citep{cohen2019certified}. \citet{Carlini2022} found that plugging diffusion models as a denoiser into the denoised smoothing framework~\citep{salman2020denoised} can lead to non-trivial certified robustness. \citet{xiao2023dense} further developed this method and studied the improvement in certified robustness.

\paragraph{Empirical evaluation of the robustness in diffusion models} Such randomness has raised concerns about gradient masking in robustness evaluation~\citep{papernot2017practical}, which may provide a false sense of robustness against gradient-based attacks~\citep{tramer2018ensemble}. \citet{athalye2018obfuscated} further identified that randomness could cause gradient masking as ``stochastic gradients'', and proposed the expectation-over-transformation (EOT) which became the standard evaluation for stochastic gradients~\citep{carlini2019evaluating}. Additionally, under the assumption that purification systems bring adversarial examples close to clean data, Backward Pass Differentiable Approximation (BPDA)~\citep{athalye2018obfuscated} was introduced as a method for evaluating purification-based defenses. However, the proper treatment of randomness in robustness evaluation remains a subject of debate~\citep{gao2022limitations,yoon2021adversarial}.

In diffusion models, internal randomness and the substantial computational overhead of full-gradient computation make robustness evaluation particularly difficult. The original DiffPure paper applied AutoAttack~\citep{croce2020reliable} with augmented SDE-based gradient estimation and reported a robust accuracy of 70.64\% on CIFAR-10. However, through a comprehensive experimental evaluation, \citep{lee2023robust} found that the robustness improvements from diffusion models were over-estimated. They recommended using the PGD-EOT with full gradients directly, and estimated the robustness around 46.84\%. \citet{li2024adbm} also challenged the original evaluation and reported a comparable robustness estimate of 45.83\%. Recently, \citet{liu2025towards} proposed to evaluate the robustness under a deterministic white box setting (DW-box), discovering that the robustness of diffusion models further decreases to 16.8\% after controlling the stochasticity.

\paragraph{Notations and preliminaries} Denote $\boldsymbol{x}_0$ as the clean image, and $\boldsymbol{x}$ as its perturbed version, so that 
\begin{equation}  
	\boldsymbol{x} = \boldsymbol{x}_0 + \epsilon \boldsymbol{\eta},  
\end{equation}  
where $\boldsymbol{\eta}$ is the normalized adversarial perturbation, and $\epsilon$ controls the magnitude of the attack.
Further denote $f$ as the purification system and $g$ as the readout classifier. Adversarial purification typically consists of two steps: (i) purifying the perturbed image using $f$; (ii) classifying the output using $g$:  
\begin{equation}  
	\hat{\boldsymbol{x}} = f(\boldsymbol{x}), \quad \boldsymbol{y} = g(\hat{\boldsymbol{x}}).  
\end{equation}  
Importantly, the purification system may be stochastic. In particular, this is true for diffusion-model-based purification.
Diffusion models consist of forward diffusion and reverse denoising processes. The forward process of Denoising Diffusion Probabilistic Models (DDPM)~\citep{ho2020denoising} is
\begin{equation}
	\label{eq:ddpm-forward}
	\bm{x}_t=f_t^{\textrm{FWD}}(\bm{x}_{t-1},\bm{\epsilon}_t)=\sqrt{\alpha_t}\bm{x}_{t-1}+\sqrt{1-\alpha_t}\bm{\epsilon}_t,
\end{equation}
in which the $\bm{\epsilon}_t\sim\mathcal{N}(\bm{0},\bm{I})$ will introduce randomness. Further, the reverse process
\begin{equation}
	\begin{aligned}
		\label{eq:ddpm-reverse}
		\bm{x}_{t-1}&=f_t^{\textrm{REV}}(\bm{x}_t,\bm{z}_t)\\
		&=\frac{1}{\sqrt{\alpha_t}}\left(\bm{x}_t-\frac{1-\alpha_t}{\sqrt{1-\bar{\alpha}_t}}\bm{\epsilon}_\theta(\bm{x}_t,t)\right)+\sigma_t\bm{z}_t,
	\end{aligned}
\end{equation}
also introduces randomness through $\bm{z}_t\sim\mathcal{N}(\bm{0},\bm{I})$. Notably, deterministic reverse process has been proposed, \textit{e.g.,} in Denoising Diffusion Implicit Models (DDIM)~\citep{song2020denoising}, the reverse process
\begin{equation}
	\begin{aligned}
		\label{eq:ddim-reverse}
		\boldsymbol{x}_{t-1}=f_t^{\textrm{DDIM}}(\bm{x}_t)&=\sqrt{\bar{\alpha}_{t-1}}\hat{\boldsymbol{x}}_0+\sqrt{1-\bar{\alpha}_{t-1}}\boldsymbol{\epsilon}_\theta(\boldsymbol{x}_t,t),\\
		\hat{\boldsymbol{x}}_0&=\frac{\boldsymbol{x}_t-\sqrt{1-\bar{\alpha}_t}\boldsymbol{\epsilon}_\theta(\boldsymbol{x}_t,t)}{\sqrt{\bar{\alpha}_t}}
	\end{aligned}
\end{equation}
is fully deterministic and thus does not involve randomness.

Denote ${\boldsymbol{\xi}}$  as a \textit{randomness configuration}, which consists the series of random noises governing the stochastic process, i.e., for DDPM, $\bm{\xi}^{\textrm{DDPM}}=
\{\bm{\epsilon}_1 \ldots \bm{\epsilon}_{t^*},\bm{z}_{t^*} \ldots \bm{z}_1 \}$. Denote $f_{\boldsymbol{\xi}}=f(\bm{x}|\bm{\xi})$ as the \textit{deterministic }purification conditioned on a particular randomness configuration ${\boldsymbol{\xi}}$.

\section{Intriguing Behaviors of Diffusion Models in Adversarial Purification}
\label{sec:hypothesis}
While the precise mechanisms underlying robustness improvements from diffusion models remain unclear, prior work has proposed intuitive explanations for how diffusion models may improve robustness. Below we empirically test two such hypotheses.
\paragraph{``Clean image attraction'' hypothesis} This hypothesis proposes that diffusion models improve robustness by purifying an adversarially perturbed image to be “closer” to the clean image, i.e.,
\begin{equation}
	\|f(\bm{x_0}+\epsilon\bm{\eta})-x_0\|\leq\epsilon\|\bm{\eta}\|.
\end{equation}
The original DiffPure paper~\citep{nie2022diffusion} indicated this view, noting that \textit{``we observe the purified images match the clean images''} and \textit{``recovering clean images from the adversarial examples''}. Intuitively, if a purification system consistently reduces $\ell_p$ distances to the clean image, it effectively transforms an adversarial perturbation into one of smaller magnitude, thereby enhancing robustness.

\begin{figure*}[ht]
	\centering
	\includegraphics[width=0.75\linewidth]{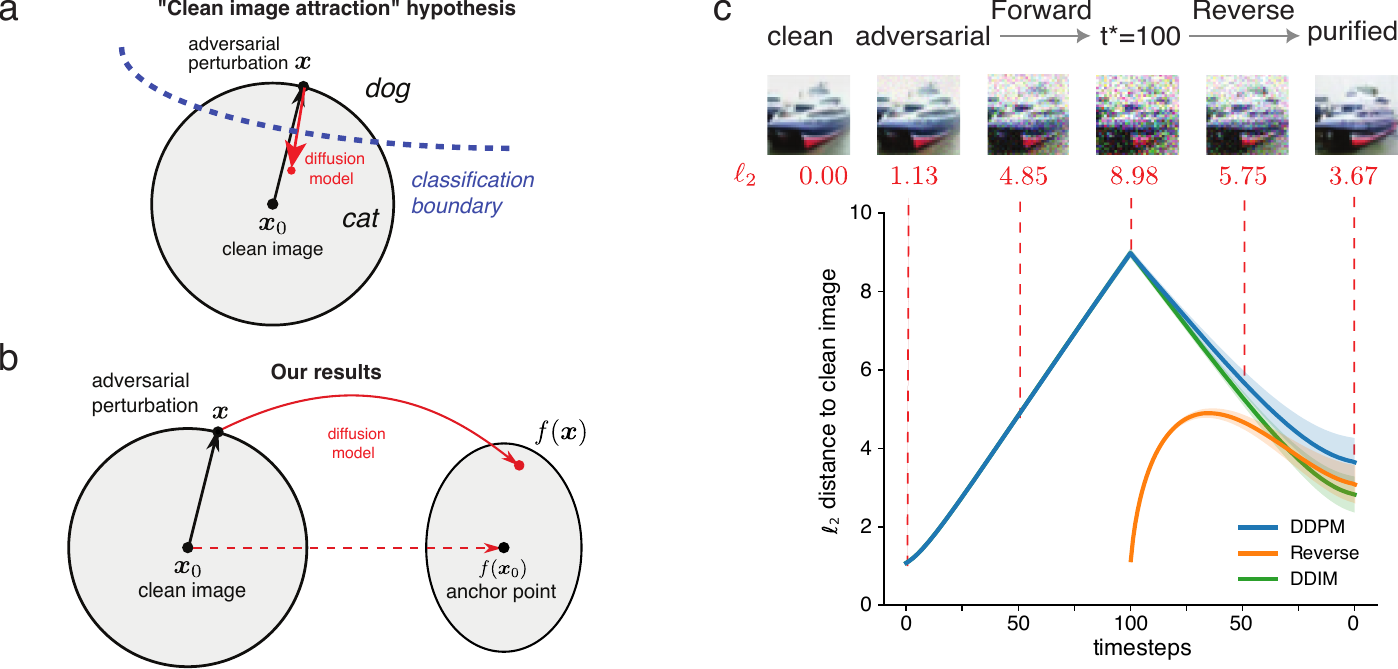}
	\caption{\textbf{Diffusion models purify states away from the clean images.} (a) Schematic showing a common hypothesis that diffusion models improve robustness by ``denoising'' inputs toward the clean image. (b) Summary of our findings, which challenge the denoising hypothesis. (c) Measured $\ell_2$ distances to clean images on CIFAR-10 during purification. We track the distances between intermediate purified states and clean images, using PGD attacks ($\ell_\infty=8/255$) as initialization. Across all methods, the purified outputs are consistently farther away from the clean image. Additional examples of adversarial and purified images can be found in Appendix~E.}
	\label{fig:anchor}
\end{figure*}

\paragraph{``Distribution alignment'' hypothesis} A more sophisticated hypothesis is that, after purification, the distribution of perturbed images becomes more closely aligned with the clean image distribution,
\begin{equation}
	D_{\textrm{KL}}[p(f(\bm{x})), p(\bm{x}_0)]\leq D_{\textrm{KL}}[p(\bm{x}),p(\bm{x}_0)],
\end{equation}
thereby improving robustness. This hypothesis is reflected in Theorem 3.1 of~\citet{nie2022diffusion}. It implies a monotonic relationship between distributional distance and robustness—distributions with smaller KL divergence from the clean distribution are expected to yield higher robustness. 

\subsection{Diffusion Models Purify Images Away from Clean Images}
\label{sec:anchor}
\paragraph{Diffusion models increase $\ell_p$ distances to clean images after purification}
To test whether diffusion models reduce the distance between adversarial and clean images, we conducted a series of experiments. From a clean image, we generated an adversarial example, then applied adversarial purification via diffusion models, and finally measured the distance between the purified image and the original clean image. Surprisingly, we found that the $\ell_2$ distance to the clean sample increased after purification (Fig.~\ref{fig:anchor}c). This phenomenon was consistent across a wide range of settings, including different attack types (BPDA, BPDA-EOT, PGD, PGD-EOT), distance metrics ($\ell_2$, $\ell_\infty$), sampling methods (DDPM, Reverse-only, DDIM) and datasets (CIFAR-10 and ImageNet). It is not specific to adversarial attacks and also holds for perturbations with uniform noise.

\begin{table}[htbp]
	\caption{Distances pre/post-diffusion models on CIFAR-10 (PGD, $\ell_\infty=8/255$).}
	\label{tab:dist-cifar10}
	\centering
	\begin{tabular}{l|cccccc}
		\toprule[1.5pt]
		{\textbf{Sampling}} & {\textbf{\bm{$\ell_2$} ($\downarrow$)}} & \textbf{\bm{$\ell_\infty$} ($\downarrow$)} & \textbf{SSIM ($\uparrow$)} \\
		\midrule
		{DDPM}  & 1.077 $\to$ 3.641 & 0.031 $\to$ 0.316  & 0.965 $\to$ 0.796 \\
		{Reverse}  & 1.149 $\to$ 3.078 & 0.031 $\to$ 0.270 & 0.965 $\to$ 0.837 \\
		{DDIM}  & 1.080 $\to$ 2.810 & 0.031 $\to$ 0.242 & 0.964 $\to$ 0.869 \\
		\bottomrule[1.5pt]
	\end{tabular}
\end{table}

\paragraph{Diffusion-based purification leads to perceptually dissimilar outputs}
While $\ell_p$ distances are highly relevant distance metrics (especially since adversarial attacks are typically defined within bounded $\ell_p$ balls), these metrics may not capture perceptual similarity. For example, translating an image by a single pixel can yield a large $\ell_2$ difference while remaining perceptually identical. Thus, we next ask if diffusion models produce outputs that are perceptually closer to the clean image, even if $\ell_p$ distances increase. To investigate this, we evaluated the structural similarity index measure (SSIM)~\citep{wang2004image}, a popular metric used in computer vision for quantifying perceptual similarity of images. As shown in Table~\ref{tab:dist-cifar10}, we observed a substantial decrease in SSIM between purified and clean images. This indicates that the purified images are not only farther away in $\ell_p$ distances, but also perceptually more dissimilar than the initial adversarial perturbations.

\subsection{Distributional Distance Fails to Explain Robustness Improvements}
We next turn to distribution-level comparisons using the Fréchet Inception Distance (FID)~\citep{heusel2017gans}. The FID score has been widely used to quantify the performance of generative models such as diffusion models. Here, we measure FID between the adversarial dataset and the clean dataset, both before and after purification, to quantify whether diffusion models bring the distribution of adversarial images closer to that of the clean samples.
Interestingly, we observe that purification with diffusion models leads to a reduction of the FID score between adversarial and clean distributions. This is consistent with the idea that diffusion models may bring the distribution of adversarial images closer to the clean data distribution \citep{li2024adbm,nie2022diffusion}.

\paragraph{Non-monotonic relation between distributional distances and adversarial robustness}
We argue that FID is not a reliable indicator of robustness. To illustrate this, we measured the FID distance between the purified adversarial samples and the clean samples across multiple purification methods and timesteps. As shown in Fig.~S4a, there is no consistently monotonic relationship between distributional distances and adversarial robustness, suggesting that the distributional alignment hypothesis alone cannot fully explain the observed robustness gains. Distance in the semantic space also fails to explain robustness improvements (Appendix~D-C).

Taken together, our empirical results challenge the hypothesis that diffusion models improve robustness by pushing adversarial images closer to their original clean images, either in the $\ell_p$ sense or in the distributional sense. 

\subsection{Diffusion Models Compress the Image Space with Fixed Randomness}
\label{sec:compress-fix}
\begin{figure*}[th]
	\centering
	\includegraphics[width=0.99\linewidth]{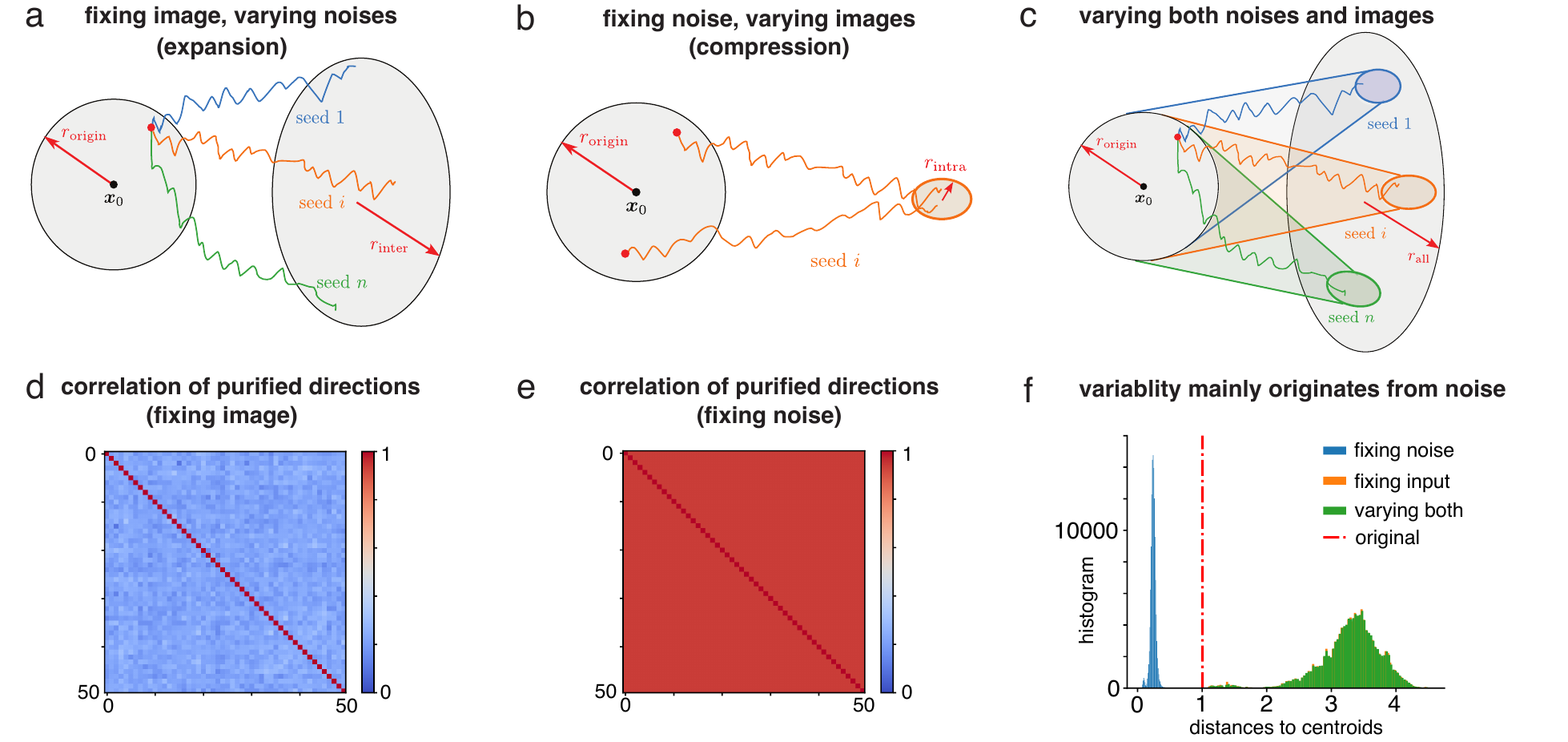}
	\caption{\textbf{Behaviors under stochasticity in diffusion models.}
		(a--c) Schematics illustrating how diffusion models transform input perturbations under different sources of variability. 
		(a) When the image is fixed and internal noise varies, purification exhibits an expansion of the input space. 
		(b) When the noise is fixed and the image varies, the input space is compressed toward a shared direction.
		(c) When both image and noise vary, internal randomness dominates, producing an overall expansion. 
		(d) Purification directions under different noise samples for the same image are weakly aligned (mean correlation: 0.22$\pm$0.003). 
		(e) Under fixed noise, purification directions across perturbed images are highly consistent (mean correlation: 0.93$\pm$0.0002). 
		(f) Distribution of $\ell_2$ distances to the centroid. Fixed-noise purification compresses the input ball (radius: 1.004 $\to$ 0.241); varying noise leads to expansion (radius: 1.004 $\to$ 3.282), confirming that internal randomness drives the dominant effect. Note that fixing the input (orange) yields almost the same distribution as varying both the input and internal randomness (green), confirming that the variance in diffusion purification is dominated by internal randomness rather than input perturbations.}
	\label{fig:random}
\end{figure*}

Since diffusion models are inherently stochastic, it is useful to investigate how much this intrinsic noise affects the output of diffusion models. Specifically,  the variability of the output of diffusion models in adversarial purification arises from two distinct sources: (i) the variability in the input images from perturbations $\boldsymbol{\eta}$ (causing variability around the anchor point), and (ii) the internal variability inherent to the purification system (causing variability of the anchor point).

Fig.~\ref{fig:random} illustrates the basic ideas. First, when fixing the input image while allowing the noise in diffusion models to vary, the output variability is solely due to the internal variability of the purification system (Fig.~\ref{fig:random}a). Surprisingly, we find that this variability is large, as indicated by the relatively low correlations of different purification directions induced by noise (Fig.~\ref{fig:random}d; mean = 0.22$\pm$0.003). Second, when fixing noise in diffusion models and allowing input images to vary (different samples from the image neighborhood), the output variability is solely induced by the variability in the input images (Fig.~\ref{fig:random}b; mean = 0.93$\pm$0.0002). Interestingly, this variability is rather small, as demonstrated in the high correlations between the purified directions 
(Fig.~\ref{fig:random}e). 
By treating input variability as the signal and internal randomness as the noise, we define a signal-to-noise ratio (SNR) of the diffusion purification (Appendix~A-D):
\begin{equation}
\text{SNR} = 
\frac{\mathbb{E}_{{\boldsymbol{\xi}}} \left[ \mathrm{Var}_{\boldsymbol{x}}f(\boldsymbol{x}|{\boldsymbol{\xi}}) \right]}
{ \mathrm{Var}_{{\boldsymbol{\xi}}} \left[ \mathbb{E}_{\boldsymbol{x}}f(\boldsymbol{x}|{\boldsymbol{\xi}}) \right]}.
\label{eq:snr}
\end{equation}
Numerically evaluating the SNR of diffusion models based on image neighborhood consistent with adversarial attack, we find the SNR is extremely low, i.e., 5.93$\pm$1.07$\times10^{-3}$, indicating
that the effect of internal stochasticity is approximately 170 times larger than that induced by input variability.

To investigate these effects further, we measured the $\ell_2$ distances between purified samples and their respective centroids. When starting from the same clean image, internal randomness leads to an expansion effect, where the $\ell_2$ radius of the perturbation space increases from 1.004$\pm$0.001 to 3.282$\pm$0.453 after purification. Furthermore, the histogram of $\ell_2$ distances shown in Fig.~\ref{fig:random}f illustrates that varying both the image and noise yields nearly the same distance distribution as varying only the noise.
Importantly, we observed the \textit{opposite effect} when randomness in diffusion models is fixed. In this case, the purified outputs become tightly clustered, indicating a strong \textit{compression effect}: the $\ell_2$ radius shrinks to 0.241$\pm$0.032.

\section{Explaining the robustness improvement in Diffusion Models}
\label{sec:randomness}
The results in Sec.~\ref{sec:compress-fix} suggest that internal 
randomness in diffusion models plays a dominant role in determining 
the final purified output, which implies that randomness may 
significantly influence the empirical evaluation of adversarial 
robustness. How to properly handle randomness in robustness evaluation 
has been a longstanding debate~\citep{athalye2018obfuscated, 
carlini2019evaluating, gao2022limitations, yoon2021adversarial}. We propose that empirically observed adversarial robustness arises from 
two distinct sources: (i) \textit{bona fide} robustness improvements 
that persist even when randomness is fixed; and (ii) the increased 
difficulty of finding effective attacks in the presence of stochasticity 
(i.e., gradient masking due to randomness). Formally, we find that the empirical 
robustness of a stochastic system $f$ can be decomposed as
\begin{equation}
\textrm{Robust}[f] = \underbrace{\textrm{Robust}[f_{\boldsymbol{\xi}
		_{\textrm{test}}}]}_{\text{\textit{bona fide} robustness}} + 
\underbrace{\textrm{Residual}[\boldsymbol{\xi}_{\textrm{test}},
	\boldsymbol{\xi}_{\textrm{attack}}]}_{\text{randomness residual}},
\label{eq:robust-decomp}
\end{equation}
where $\boldsymbol{\xi}_{\textrm{test}}$ denotes the randomness 
configuration at the time of evaluation, and 
$\boldsymbol{\xi}_{\textrm{attack}}$ denotes the randomness 
used during attack generation. The \textit{bona fide} 
robustness is defined as the robustness when the randomness is 
fixed as the same during attack and evaluation, i.e., 
$\boldsymbol{\xi}_{\textrm{attack}}=\boldsymbol{\xi}_{\textrm{test}}$ 
(Sec.~\ref{sec:random-fix}); the remainder term, which we 
define as the \textit{randomness residual}, is determined by 
the cosine similarity between the perturbations generated under 
$\boldsymbol{\xi}_{\textrm{attack}}$ and 
$\boldsymbol{\xi}_{\textrm{test}}$ (Sec.~\ref{sec:random-residual}). 

As we show below, the first term in Eq.~\ref{eq:robust-decomp}, $\textrm{Robust}[f_{\boldsymbol{\xi}_{\textrm{test}}}]$, is determined by how much diffusion models compress the image neighborhood. The second term,  $\textrm{Residual}[\boldsymbol{\xi}_{\textrm{test}},
	\boldsymbol{\xi}_{\textrm{attack}}]$, is determined by the cosine similarity of the actual attack and the optimal attack (when randomness is fixed). For a given diffusion-based defense, its robustness is determined by a combination of these two factors. In this way, our results provide a unifying and principled account of the observed robustness under various diffusion-based purification methods.

\subsection{EOT as a Transfer Attack}
We will first show that the traditional evaluation method for dealing with stochasticity, Expectation-over-Transformations (EOT)~\citep{athalye2018obfuscated}, is not sufficient, and should be interpreted as a transfer attack.

Following the notation in Sec.~\ref{sec:notation}, let $L$ be the 
loss function and $t$ be the target label associated with 
$\boldsymbol{x}_0$. The attack gradient and the test-time system 
can be expressed as
\begin{equation}
\textrm{Attack:}\;\;
\nabla_{\boldsymbol{x}}L\left[g(f_{{\boldsymbol{\xi}}_{\textrm{attack}}}
(\boldsymbol{x})), t\right],
\qquad\textrm{Test:}\;\;
g(f_{{\boldsymbol{\xi}}_{\textrm{test}}}(\boldsymbol{x})).
\label{eq:attack-random}
\end{equation}
A critical observation is that ${\boldsymbol{\xi}}_{\textrm{attack}}$ 
and ${\boldsymbol{\xi}}_{\textrm{test}}$ are independently sampled 
from the same distribution, and therefore typically differ. 
Consequently, the attack optimizes against a \emph{different} 
instantiation of the defense than the one encountered at test time---a 
setting that is precisely the definition of a transfer attack. 
This mismatch is especially severe for diffusion models, where 
internal stochasticity heavily determines the purified output 
(Sec.~\ref{sec:compress-fix}), making the discrepancy between 
$f_{{\boldsymbol{\xi}}_{\textrm{attack}}}$ and 
$f_{{\boldsymbol{\xi}}_{\textrm{test}}}$ substantial.

Expectation-over-Transformations (EOT)~\citep{athalye2018obfuscated} 
was proposed to address this suboptimality by marginalizing over 
randomness during gradient computation:
\begin{equation}
\begin{aligned}
	\textrm{EOT:}\quad
	&\mathbb{E}_{\boldsymbol{\xi}}\nabla_{\boldsymbol{x}}L[g(f_{{\boldsymbol{\xi}}
		_{\textrm{attack}}}(\boldsymbol{x})), t]
	=\nabla_{\boldsymbol{x}}\mathbb{E}_{\boldsymbol{\xi}} 
	L[g(f_{{\boldsymbol{\xi}}_{\textrm{attack}}}(\boldsymbol{x})), t], \\
	\textrm{Test:}\quad
	&g(f_{{\boldsymbol{\xi}}_{\textrm{test}}}(\boldsymbol{x})).
\end{aligned}
\end{equation}
Crucially, while EOT reduces gradient variance during the attack, 
randomness \emph{still remains} at \textit{the test time}. Concretely, 
PGD-EOT approximates the gradient of the \emph{ensemble} system 
$\mathbb{E}_{\boldsymbol{\xi}} g(f_{{\boldsymbol{\xi}}}(\boldsymbol{x}))$, 
but then applies the resulting perturbation to the 
\emph{non-ensemble} system $g(f_{{\boldsymbol{\xi}}_{\textrm{test}}}
(\boldsymbol{x}))$. Under the approximation that expectation 
commutes with the loss,\footnote{This does not hold exactly for 
cross-entropy loss, a subtlety not explicitly discussed 
in~\citet{athalye2018obfuscated}.} this gives:
\begin{equation}
\textrm{Attack (EOT):}\quad
\nabla_{\boldsymbol{x}} L\!\left[\mathbb{E}_{\boldsymbol{\xi}}\, 
g(f_{{\boldsymbol{\xi}}}(\boldsymbol{x})),\, t\right].
\end{equation}
This formulation implies that PGD-EOT should be interpreted as a transfer attack: 
it optimizes against the ensemble system, yet evaluates against 
a single random instantiation. While EOT yields better gradient 
estimates than single-sample attacks---as we confirm empirically---the 
mismatch between the EOT direction and optimal attack direction still exists.

We quantify the mismatch by measuring the correlation 
between PGD-EOT gradients and the oracle gradients that directly 
target $g(f_{{\boldsymbol{\xi}}_{\textrm{test}}}(\boldsymbol{x}))$, 
finding a mean correlation of $0.1682 \pm 0.0976$ 
(Table~\ref{tab:corr-attack})---confirming that EOT gradients 
deviated substantially from the optimal gradient.

To empirically test our theoretical reasoning, we perform transfer attack experiments across different randomness configurations. 
Specifically, we generate adversarial perturbations under one fixed 
randomness configuration and evaluate them under an independently
sampled fixed configuration---directly mimicking the attack--test 
mismatch described above. We hypothesize that this procedure should 
reproduce the robustness numbers observed under uncontrolled 
randomness setting. In support of this hypothesis, the resulting robustness  on both CIFAR-10 and ImageNet is consistent with 
previously reported numbers (Tables~\ref{tab:res-transfer} 
and~\ref{tab:res-cifar10}).
To further quantify the degree of mismatch, we measure the 
correlation between adversarial perturbations generated under 
different randomness configurations, finding a mean correlation 
of only $0.0818 \pm 0.0709$ (Table~\ref{tab:corr-attack}). 
This low correlation confirms that stochasticity in diffusion 
models severely hinders the attacker from finding perturbations 
that are effective at test time. 

Taken together, EOT partially 
alleviates this issue of sub-optimal attack by increasing the cross-configuration 
correlation from $0.0818$ to $0.1682$, however, it remains far from 
optimal. Crucially, the EOT gradient correlation is 
only marginally higher than the cross-configuration transfer 
correlation, suggesting that EOT more closely resembles a transfer attack across randomness configurations than a faithful approximation of the true gradient.

\begin{table}[htbp]
\centering
\caption{Transfer attack across different randomness.}
\label{tab:res-transfer}
\centering
\begin{tabular}{lc|c}
	\toprule[1.5pt]
	{\textbf{Dataset}} & {\textbf{Fix Random}} & {\textbf{PGD/BPDA}} \\
	\midrule
	CIFAR-10 & \cmark & 77.4$\pm$0.36\% \\
	ImageNet & \cmark & 66.3$\pm$2.32\% \\
	\bottomrule[1.5pt]
\end{tabular}
\vspace{5mm}
\caption{Correlations between PGD/EOT attacks on CIFAR-10.}
\label{tab:corr-attack}
\centering
\begin{tabular}{l|c}
	\toprule[1.5pt]
	{\textbf{Attacks}} & {\textbf{Correlation}} \\
	\midrule
	PGD (Fix, Seed 0) vs. PGD (Fix, Seed 1) & 0.0818$\pm$0.0709 \\
	PGD-EOT vs. PGD (Fix, Seed 0) & 0.1682$\pm$0.0976 \\
	\bottomrule[1.5pt]
\end{tabular}
\end{table}

\subsection{Robustness of Diffusion Models Without Stochasticity}
\label{sec:random-fix}

We next use an alternative evaluation protocol that eliminates the 
attack--test mismatch entirely by fixing the randomness across both 
attack generation and evaluation---specifically, by setting 
${\boldsymbol{\xi}}_{\textrm{attack}} = {\boldsymbol{\xi}}_{\textrm{test}}$. 
Under this protocol, the attack directly optimizes against the 
same function instantiation that will be evaluated, removing 
concerns about suboptimal gradients and transfer effects. This 
yields a faithful estimate of $\textrm{Robust}[f_{\boldsymbol{\xi}
_{\textrm{test}}}]$---the \emph{bona fide} robustness term in 
our decomposition (Eq.~\ref{eq:robust-decomp})---isolating true 
robustness from randomness-induced gradient masking.

We implement this protocol by fixing the random seeds for both 
the forward and reverse processes of the diffusion model during 
attack generation and evaluation (see Appendix~C 
for implementation details). This technique was proposed in our earlier version of this work reported in \citep{yuezhang2023dissecting}.
Note that a concurrent study~\citep{liu2025towards} independently advocates 
the same fixed-randomness evaluation strategy. Going beyond previous studies, our current work 
provides a formal decomposition into 
\textit{bona fide} robustness and randomness residual 
(Eq.~\ref{eq:robust-decomp}), analyzing both terms theoretically 
and empirically (Sec.~\ref{sec:random-residual} 
and~\ref{sec:compress}), and characterizing the geometric 
structure underlying the residual term. 

\begin{table}[htbp]
\caption{Robustness of diffusion models w/o.\ stochasticity 
	on CIFAR-10 ($\ell_\infty=8/255,\,t=100$).}
\label{tab:res-cifar10}
\centering
\begin{tabular}{lc|ccc}
	\toprule[1.5pt]
	{\textbf{Model}} & {\textbf{Fix Random}} & 
	{\textbf{Clean Acc.}} & {\textbf{PGD}} & {\textbf{PGD-EOT}} \\
	\midrule
	\multirow{2}{*}{DDPM} & \xmark & 86.0$\pm$0.8\% & 
	71.9$\pm$0.2\% & 59.3\% \\
	& \cmark & 85.8$\pm$0.4\% & \textbf{23.7$\pm$0.7\%} & -- \\
	\bottomrule[1.5pt]
\end{tabular}
\vspace{5mm}
\caption{Robustness of diffusion models w/o.\ stochasticity 
	on ImageNet ($\ell_\infty=4/255,\,t=150$).}
\label{tab:res-imagenet}
\centering
\begin{tabular}{lc|ccc}
	\toprule[1.5pt]
	{\textbf{Model}} & {\textbf{Fix Random}} & 
	{\textbf{Clean Acc.}} & {\textbf{BPDA}} & {\textbf{BPDA-EOT}} \\
	\midrule
	\multirow{2}{*}{Guided} & \xmark & 67.2$\pm$2.4\% & 
	63.7$\pm$1.2\% & 59.0\% \\
	& \cmark & 68.5$\pm$0.8\% & \textbf{29.5$\pm$0.4\%} & -- \\
	\bottomrule[1.5pt]
\end{tabular}
\end{table}

Our results show that fixing randomness reveals a dramatic reduction in the observed 
robustness (see Tables~\ref{tab:res-cifar10} and~\ref{tab:res-imagenet}.). On CIFAR-10, robust accuracy drops from $71.9\%$ 
(uncontrolled, PGD) to $23.7\%$ under fixed randomness---a 
reduction of over 48 percentage points. On ImageNet, robust 
accuracy falls from $63.7\%$ to $29.5\%$ under BPDA. These 
results confirm that the majority of previously reported 
robustness gains are attributable to the residual term in 
Eq.~\ref{eq:robust-decomp}, i.e., gradient masking induced 
by stochasticity, rather than bona fide robustness of the 
purification system. Importantly, even under fully controlled 
randomness, diffusion models retain non-trivial robustness 
($23.7\%$ on CIFAR-10, $29.5\%$ on ImageNet), suggesting 
that the purification process does provide genuine (albeit 
substantially smaller than what was initially proposed) protection against adversarial attacks.

\begin{figure*}[th]
\centering
\includegraphics[width=0.96\linewidth]{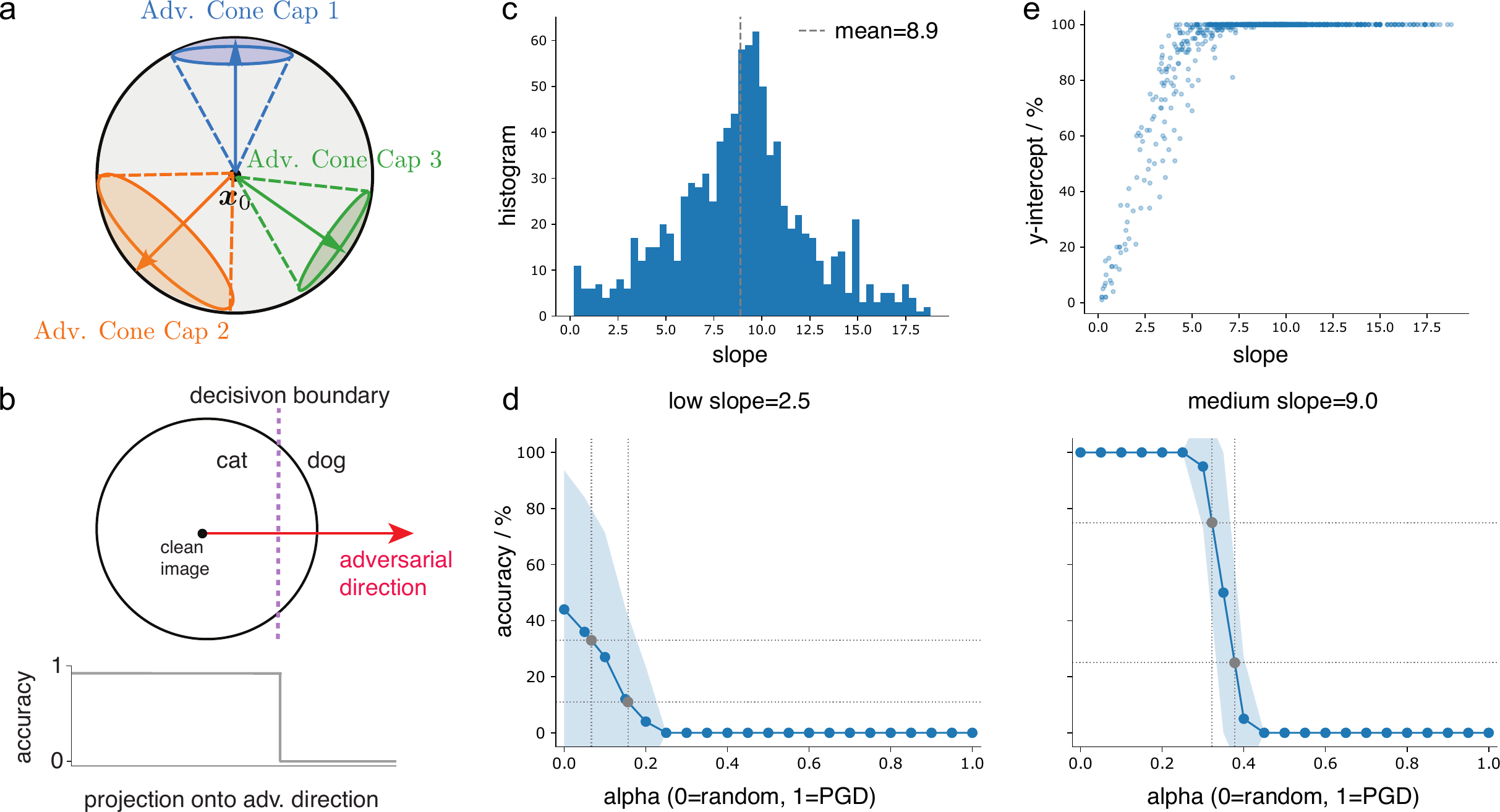}
\caption{\textbf{The hyperspherical cap model of adversarial regions.}
	(a) Schematic of the hyperspherical cap model.
	(b) A key prediction of the hyperspherical cap model is that the transition of the categorical decision after projecting onto the adversarial direction should be sharp. 
	(c) Histogram of the transition slope across 1000 
	CIFAR-10 test images. The distribution is approximately 
	bell-shaped with a mean slope of 8.9. (d) Representative transition curves with a low slope ($2.5$) and a medium slope ($9.0$). The medium-slope curve represents the mean of the distribution and constitutes a visually sharp transition. (e) The y-intercept of the transition curve is 
	positively correlated with the slope. The y-intercept represents classifier accuracy under small random perturbations 
	at the same scale as the adversarial budget. This plot reveals 
	that the dataset intrinsically contains a small portion
	of samples residing close to the decision boundary--susceptible 
	to random perturbations without adversarial optimization--yielding low y-intercepts and shallower transitions. The 
	remaining majority lie away from the boundary, producing 
	high y-intercepts and steep transitions consistent with the 
	hyperspherical cap model.}
\label{fig:hypercap}
\end{figure*}

\subsection{Randomness Residual and the Hyperspherical Cap Model}
\label{sec:random-residual}

Having established that fixing randomness dramatically reduces 
observed robustness, we now turn to characterizing the geometric 
structure underlying the residual term 
$\textrm{residual}[\boldsymbol{\xi}_{\textrm{test}}, 
\boldsymbol{\xi}_{\textrm{attack}}]$. Specifically, we ask: 
\emph{what determines how large this residual robustness is for a given 
attack?}

\paragraph{The hyperspherical-cap model of adversarial regions}
To answer this question, we first develop a hyperspherical-cap model of
the adversarial region of deep neural networks in the neighborhood of a given clean image.
Consider a
clean input $\boldsymbol{x}_0$ correctly classified by a base
classifier $g$ within an $\ell_2$-ball of radius $\epsilon$,
and let $\boldsymbol{\eta}_{\textrm{adv}}$ be an
\emph{adversarial direction}---one along which a single logit
changes far more than along a random direction. We will make two assumptions: (i) \emph{sparsity}: adversarial
directions are rare, so a generic direction orthogonal to
$\boldsymbol{\eta}_{\textrm{adv}}$ does not appreciably change
the logits; and (ii) \emph{critical threshold}: along
$\boldsymbol{\eta}_{\textrm{adv}}$, the decision boundary is
crossed exactly once, at radius $\gamma\le\epsilon$. Under
these assumptions, when $\epsilon$ is small the local geometry
is well approximated by linearization, and a perturbed point
$\boldsymbol{x}_0+\boldsymbol{\eta}$ is misclassified iff the
component of $\boldsymbol{\eta}$ along
$\boldsymbol{\eta}_{\textrm{adv}}$ exceeds $\gamma$. The set
of unit-norm adversarial directions therefore forms a
\emph{hyperspherical cap}---the intersection of the unit
hypersphere with a half-space whose normal is
$\boldsymbol{\eta}_{\textrm{adv}}$, or equivalently the
intersection of the perturbation hypersphere with a hypercone
whose axis is $\boldsymbol{\eta}_{\textrm{adv}}$
(Fig.~\ref{fig:hypercap}a). Multiple disjoint adversarial caps
may coexist when distinct adversarial directions are present.
A formal definition, the two assumptions, and the derivation
are given in Appendix~B. We find that the proposed hyperspherical cap model can be used to prove the GAAS (Gradient Aligned Adversarial Subspaces) algorithm proposed in \cite{tramer2017space} that quantifies the number of orthogonal directions that can be fit into the adversarial region (see Appendix~B). 

\paragraph{Sharp decision transitions support the hyperspherical cap model}
The hyperspherical cap model leads to a testable prediction: as a perturbation's
direction is rotated from a random direction (which, by
sparsity, lies outside the cap with high probability) toward
the adversarial direction of a deep network (the cap axis), classifier accuracy
should drop abruptly once the cap boundary is crossed
(Fig.~\ref{fig:hypercap}b). We test this prediction directly using the base
CIFAR-10 classifier, which has no internal randomness. For
each of $1000$ test images we (i) compute the adversarial
direction $\boldsymbol{\eta}_{\textrm{adv}}$ via PGD on the
base classifier, (ii) sample $100$ random directions
$\boldsymbol{\eta}_{\textrm{random}}$ at the same $\ell_2$
budget, and (iii) for each random direction interpolate
$\hat{\boldsymbol{\eta}}(\alpha)\propto
\alpha\cdot\boldsymbol{\eta}_{\textrm{adv}}+(1-\alpha)
\cdot\boldsymbol{\eta}_{\textrm{random}}$ with
$\alpha\in[0,1]$, normalized to constant $\ell_2$ norm. This
is the same $\alpha$-interpolation construction we use for
the residual analysis below. Averaging classifier accuracy
across the $100$ random directions yields a transition curve
per image, and we summarize its steepness by the
$25$--$75\%$ \emph{slope}---the inverse $\alpha$-distance
over which accuracy falls from $75\%$ to $25\%$. Across the
dataset the slopes are bell-shaped with mean $8.9$
(Fig.~\ref{fig:hypercap}c), corresponding to a visually sharp
transition (Fig.~\ref{fig:hypercap}d, right). The slope is
further positively correlated with the curve's
y-intercept---classifier accuracy under random perturbations
at the adversarial budget (Fig.~\ref{fig:hypercap}e). This
correlation reveals two regimes: a small subset of inputs
lies close to the decision boundary, vulnerable to even
random perturbations, and yields shallow transitions with low
y-intercepts; the majority lie well inside their decision
region and produce the steep transitions predicted by the
hyperspherical-cap model.

\paragraph{Hyperspherical cap geometry explains the magnitude of randomness residual}
The empirical sharpness measures of the transition reported above (Fig.~\ref{fig:hypercap}) validate the cap
model's central assumption, that is, locally adversarial
vulnerability depends primarily on \emph{angular alignment}
with the adversarial direction, with negligible structure in
orthogonal directions. We now apply this insight to understand the
fixed-randomness diffusion system $f_{\boldsymbol{\xi}_{
	\textrm{test}}}$, which is a deterministic function once
$\boldsymbol{\xi}_{\textrm{test}}$ is fixed and therefore
admits its own adversarial direction
$\boldsymbol{\eta}(\boldsymbol{\xi}_{\textrm{test}})$
(obtained by PGD-fix). Under the cap model, an attack
$\boldsymbol{\eta}(\boldsymbol{\xi}_{\textrm{attack}})$
generated under a different randomness configuration is
effective on $f_{\boldsymbol{\xi}_{\textrm{test}}}$ to the
extent that it falls inside the test-time cap---that is, to
the extent that its angular correlation with
$\boldsymbol{\eta}(\boldsymbol{\xi}_{\textrm{test}})$ is high.
The residual robustness (randomness residual) should therefore decrease monotonically with this
correlation, with the locally linear cap geometry setting a
first-order baseline. 

\begin{figure}[t]
\centering
\includegraphics[width=0.85\linewidth]
{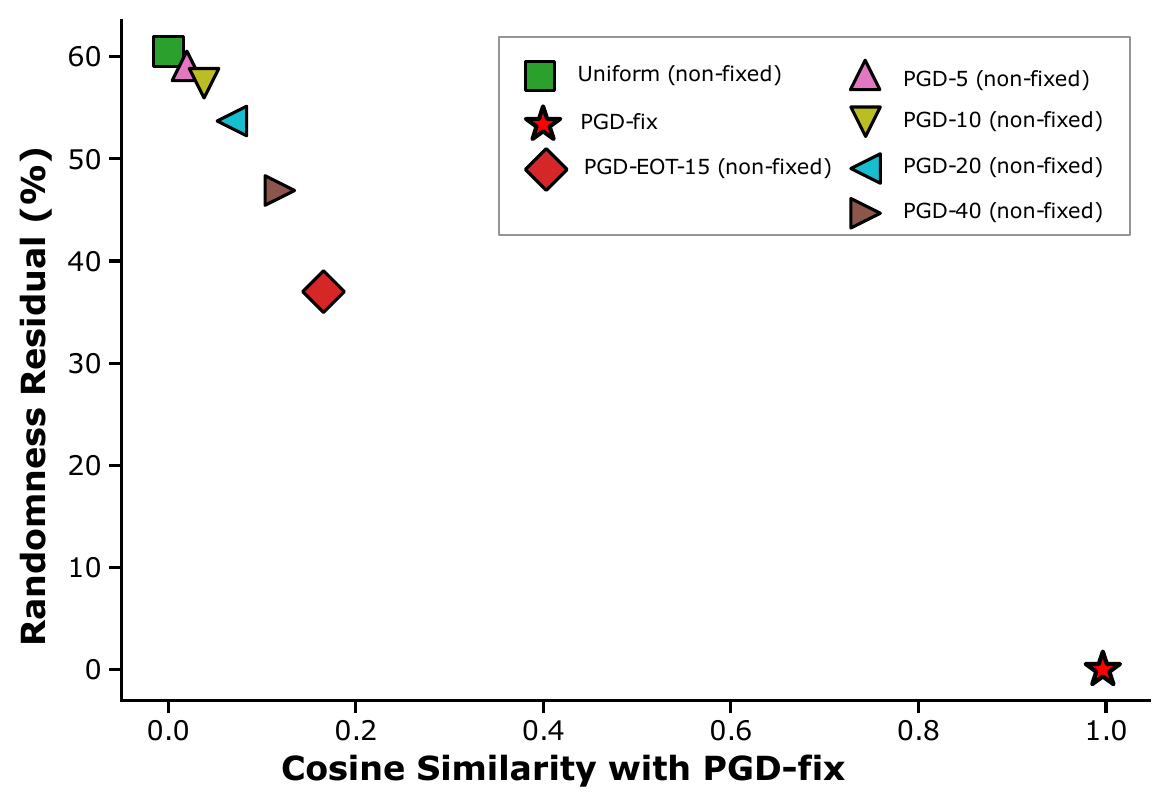}
\caption{
	\textbf{Empirical evaluation reveals a systematic relation between randomness residual (\textit{i.e.,} robustness gain due to randomness during the attack) and the cosine similarity of the actual vs. fixed-randomness attack 
		(PGD-fix).} 
	Colored markers show non-fixed attacks (PGD with varying 
	steps, PGD-EOT, and Uniform noise) plotted at their 
	measured correlation with 
	$\boldsymbol{\eta}(\boldsymbol{\xi}_{\textrm{test}})$. 
    PGD-fix (star) 
	serves as a reference point when correlation $= 1.0$, 
	residual $= 0$. These results reveal a systematic relationship between the randomness residual and the cosine similarity between the actual attack and PGD-fix, consistent with our theoretical prediction. 
}
\label{fig:robust-residual}
\end{figure}

A natural measure of how well an attack under randomness
configuration $\boldsymbol{\xi}_{\textrm{attack}}$ approximates
the optimal attack under $\boldsymbol{\xi}_{\textrm{test}}$ is
the cosine similarity (essentially the Pearson correlation) between the resulting perturbations:
\begin{equation}
\rho = \frac{\langle \boldsymbol{\eta}(\boldsymbol{\xi}_{
		\textrm{attack}}),\, 
	\boldsymbol{\eta}(\boldsymbol{\xi}_{\textrm{test}}) \rangle}
{\|\boldsymbol{\eta}(\boldsymbol{\xi}_{\textrm{attack}})\|_2 
	\, \|\boldsymbol{\eta}(\boldsymbol{\xi}_{\textrm{test}})\|_2},
\end{equation}
where $\boldsymbol{\eta}(\boldsymbol{\xi}_{\textrm{test}})$ 
denotes the perturbation from PGD-fix---the oracle attack 
optimized under the test randomness configuration.  A large $\rho$ implies that the attack has a large projection on the optimal attack direction, and according to our theory is more likely to break the system. 
We find 
that this correlation is strongly and monotonically dependent on the randomness residual, \textit{i.e.,} as $\rho$ increases, the randomness 
residual decreases systematically (Fig.~\ref{fig:robust-residual}), and eventually reaches 0 when $\rho =1$ (which correspond to PGD attack with fixed randomness).

We did not obtain data points for cosine similarity between 0.2 and 1, because all the attacks examined with randomness during the testing phase are only weakly correlated with PGD-fix. It is useful to note that our theory predicts that for attacks with larger cosine similarity with PGD-fix (one possibility would be EOT-based attack based on many more samples), the empirically obtained robustness would be lower. Taken together, these results provide a geometric account of the randomness residual, and show that its magnitude is primarily governed by how well the attack aligns with the fixed-randomness adversarial direction.

\section{Understanding Adversarial Purification through Compression}
\label{sec:compress}
In this section, we will study the mechanisms underlying the improvement in robustness when \textit{randomness is fixed} in diffusion models. We will show that it is the compression of the image neighborhood that leads to the robustness improvement, and furthermore, the amount of adversarial robustness is tightly predicted by the amount of compression (quantified by the expected compression rate ($\overline{\textrm{CR}}$)).

\subsection{A Compression Framework of Adversarial Purification}
Following the notations in Sec.~\ref{sec:notation}, for randomness configuration ${\boldsymbol{\xi}}$, consider the Taylor expansion of $f_{\boldsymbol{\xi}}(\boldsymbol{x})$ around $\boldsymbol{x}_0$:  
\begin{equation}  
f_{\boldsymbol{\xi}}(\boldsymbol{x}) = f_{\boldsymbol{\xi}}(\boldsymbol{x}_0 + \epsilon \boldsymbol{\eta}) = \underbrace{f_{\boldsymbol{\xi}}(\boldsymbol{x}_0)}_{\textrm{anchor point}} + \epsilon \underbrace{J_{f_{\boldsymbol{\xi}}}(\boldsymbol{x}_0)}_{\textrm{compression}}\boldsymbol{\eta} + o(\epsilon),
\end{equation}
where $J_{f_{\boldsymbol{\xi}}}(\boldsymbol{x}_0)$ is the Jacobian matrix that quantifies the local linear transformation induced by $f_{\boldsymbol{\xi}}$.

\paragraph{Purified clean image $f_{\bm{\xi}}(\bm{x}_0)$ as the anchor point}We observe that the purified clean image $f_{\bm{\xi}}(\bm{x}_0)$ naturally defines the center of compression, which we refer to as the \textit{anchor point}. As discussed in Sec.~\ref{sec:hypothesis}, in contrast to the clean image attraction hypothesis, diffusion models do not satisfy $f(\bm{x}_0)=\bm{x}_0$. This discrepancy induces a slight drop in clean accuracy, depending on the purification step $t^*$ (Tables~\ref{tab:res-cifar10}, \ref{tab:res-imagenet}). Furthermore, as analyzed in Sec.~\ref{sec:randomness}, each anchor point is uniquely determined by the specific randomness configuration $\bm{\xi}$.

\paragraph{Compression rates under infinitesimal isotropic noise} To quantify the magnitude of the compression effect, we define a scalar measure, the compression rate (CR) of a perturbation vector:
\begin{equation}
\textrm{CR}[{f_{\boldsymbol{\xi}}},\bm{x}_0,{\boldsymbol{\eta}}] = \frac{\|f_{\boldsymbol{\xi}}(\boldsymbol{x}_0+\boldsymbol{\eta}) - f_{\boldsymbol{\xi}}(\boldsymbol{x}_0)\|}{\|\boldsymbol{\eta}\|},
\end{equation}
which is evaluated for a specific perturbation $\bm{\eta}$. One can further define the expected compression rate ($\overline{\textrm{CR}}$) for perturbations sampled from infinitesimal isotropic noise (i.e., $\mathbb{E}[\bm{\eta}]=\bm{0},\,\mathbb{E}[\bm{\eta}\bm{\eta}^T]=I_d$):
\begin{equation}
\overline{\textrm{CR}}[f_{\bm{\xi}},\bm{x}_0] = \mathbb{E}_{\bm{\eta}}\textrm{CR}[{f_{\boldsymbol{\xi}}}(\boldsymbol{x}_0),\epsilon\bm{\eta}],\quad\epsilon\to0,\,\bm{\eta}\sim\textrm{Iso}(d).
\end{equation}
This quantity captures the average local compression effect of $f_{\bm{\xi}}$ around $\bm{x}_0$. In practice, we sample uniform noise $\boldsymbol{\eta}$ with the same scale as the adversarial perturbation budget $\epsilon$, thereby matching the $\ell_p$ attack setting. As we will show below, $\overline{\textrm{CR}}$ is closely connected to adversarial robustness. Importantly, because it is defined as an expectation rather than tied to a particular $\bm{\eta}$, $\overline{\textrm{CR}}$ can be estimated easily using sampling, without expensive gradient computations.

\paragraph{Relation between compression rates and robustness}
Adversarial examples arise when a small neighborhood around clean image intersects the classifier’s decision boundary. We reason that the compression of image space effectively reduces the size of image neighborhood and thus reduces the odds of the transformed neighborhood intersecting with a decision boundary. Thus, we predict that there should be a direct relationship between the compression rate and adversarial robustness.

\begin{figure*}[ht]
\centering  
\includegraphics[width=0.99\linewidth]{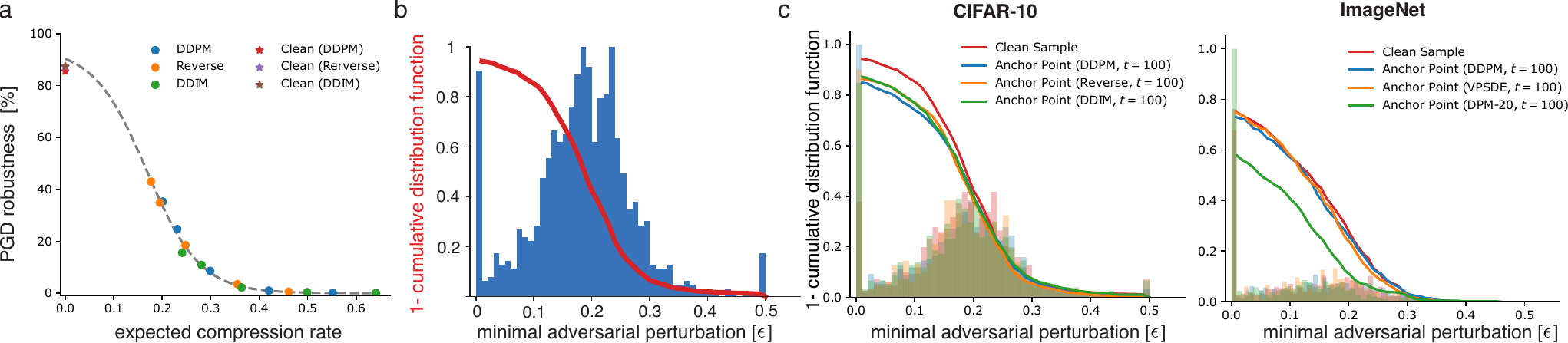}
\caption{\textbf{Lawful relation between expected compression rates and adversarial robustness.} (a) The compression rate and robustness without stochasticity of diffusion models follow a consistent relation well captured by a sigmoid function. Note that the curve generalizes across different sampling methods and extrapolates smoothly to clean accuracies at the y-intercept. (b) The sigmoidal relation arises from the survival function ($1-\textrm{CDF}$) of the distribution of minimum adversarial perturbations of the classifier, an intrinsic property of the classifier that is independent of purification systems. (c) The distribution of minimum perturbations at the anchor points and the survival functions (1--CDF). For both CIFAR-10 and ImageNet, and across the anchor points of discrete and continuous sampling methods, the minimum perturbation distributions exhibit a zero-inflated Gaussian shape qualitatively similar to that at the clean samples, where the zero mass corresponds to the clean accuracy at the anchor points.}
\label{fig:relation}
\end{figure*}

We test this predicted relation using CIFAR-10. We take advantage of the fact that different implementations of diffusion models (DDPM, Reverse-only, DDIM) lead to different adversarial robustness (Table~\ref{tab:cr-cifar10}). Examining a collection of diffusion models with various hyperparameters, we plot their robustness performances against their compression rates. Strikingly, we find that the two quantities exhibit a \textit{lawful relationship} (Fig. ~\ref{fig:relation}a). The compression-robustness curve increases sharply for a compression rate around 0.2. We find that the relation between compression and robustness can be well fitted by a sigmoidal function. Interestingly, extrapolating the fitted curve to zero compression rate leads to a relatively accurate prediction of the clean accuracy.

\begin{table}[t]
	\caption{Compression rates and adversarial robustness on CIFAR-10.}
	\label{tab:cr-cifar10}
	\centering
	\footnotesize
	\setlength{\tabcolsep}{6pt}
	\begin{tabular}{c|cc|cc|cc}
		\toprule[1.2pt]
		\multirow{2}{*}{\textbf{$t$}}
		& \multicolumn{2}{c|}{\textbf{DDPM}} 
		& \multicolumn{2}{c|}{\textbf{Reverse-only}} 
		& \multicolumn{2}{c}{\textbf{DDIM}} \\
		& $\overline{\textrm{CR}}$ & PGD-Fix 
		& $\overline{\textrm{CR}}$ & PGD-Fix
		& $\overline{\textrm{CR}}$ & PGD-Fix \\
		\midrule
		10  & 0.552 & 0.1\%  & 0.461 & 0.6\%  & 0.641 & 0.1\% \\
		20  & 0.420 & 1.0\%  & 0.355 & 3.5\%  & 0.499 & 0.4\% \\
		50  & 0.299 & 8.6\%  & 0.248 & 18.5\% & 0.364 & 2.2\% \\
		100 & 0.231 & 23.7\% & 0.195 & 34.9\% & 0.281 & 10.9\% \\
		150 & 0.201 & 35.3\% & 0.177 & 43.0\% & 0.241 & 15.6\% \\
		\bottomrule[1.2pt]
	\end{tabular}
\end{table}

\subsection{Sigmoidal Relation Arises from the Distribution of Minimum Perturbation of the Classifier}
The observed sigmoidal relationship between compression rate and robustness points to a more fundamental explanation. Building on our compression-based theory, we hypothesize that this pattern reflects the intrinsic link between attack budget and robustness. To investigate, we estimate the distribution of minimum perturbation magnitudes required to fool the CIFAR-10 classifier by performing a binary search over perturbation scales $\epsilon$ for each test image. As shown in Fig.~\ref{fig:relation}b, the resulting distribution is well described by a zero-inflated Gaussian: the Gaussian component captures the bulk of perturbation thresholds, while the zero-inflated component reflects clean input errors, for which any perturbation suffices. This characterization explains the sigmoidal robustness curve—the \textit{survival function} ($1-\textrm{CDF}$) of such a distribution naturally yields a sigmoid, with the plateau determined by the classifier’s clean accuracy. These findings suggest that the sigmoidal relation arises from the intrinsic distribution of minimum perturbations imposed by the classifier’s decision boundary, rather than from properties of the purification system itself. Consequently, different purification methods (e.g., DDPM, reverse-only, DDIM) fall on the same curve because they compress perturbations to varying degrees but are governed by the same classifier-induced distribution. Moreover, this perspective provides a principled way to predict adversarial robustness from compression rates alone, without requiring gradient-based evaluation.

Strictly speaking, this argument only holds if the distribution of the minimum perturbations of the classifier at the clean points is similar to that at the anchor points. In this section, we verify that it is indeed the case, as long as the clean accuracy at the anchor point is close to that at the clean samples. Predicting at the anchor points, would, however, always provide a slightly better estimation.

We have measured the distributions of the classifier’s minimum perturbations at the anchor points of DDPM, Reverse-only DDPM, and DDIM models on CIFAR-10, and  at the anchor points of DDPM, VPSDE, and DPM-20 on ImageNet. As shown in Fig.~\ref{fig:relation}c, in all cases, the distributions exhibit a zero-inflated Gaussian shape. Consequently, their survival functions (i.e., $1-\mathrm{CDF}$) take on a sigmoidal form. The y-intercept of the survival function, arising from the zero-mass component, corresponds precisely to the clean accuracy at the anchor point.

As a result, the sigmoidal curves at the anchor points closely match those at the clean samples whenever the clean accuracies remain similar (e.g., for $t \leq 100$). This condition is necessary for any adversarial purification procedure to be meaningful. When $t$ becomes large and the clean accuracy at the anchor points drops substantially, the two sigmoidal curves naturally diverge.

Therefore, in the regime where adversarial purification is meaningful—that is, where the clean accuracy at the anchor points does not significantly decrease relative to that of the original samples—the sigmoidal relation at the anchor points can be well approximated by that of the clean samples. This approximation greatly simplifies the analysis, as the resulting sigmoidal structure becomes an intrinsic property of the classifier itself, independent of the purification process. Incorporating the small deviations between the two distributions would further refine the quantitative precision of our theory.

\subsection{Estimating Adversarial Robustness with Compression Rates}
We next investigate whether it would be possible to go one step further by using the compression rate (CR) and the minimal adversarial perturbation of the base classifier to predict the robustness of a purification system. For a given purification system, we can compute (i) the expected CR, and (ii) the survival function for the base classifier at the anchor points (akin to red curve in Fig.~\ref{fig:relation}b). Using (ii), we identify the survival probability corresponds to the CR from (i), and use this survival probability as the predicted adversarial robustness.

We investigate how well such predictions based on CR match empirical robustness for CIFAR-10. As shown in Table~\ref{tab:cr-cifar10-pred}, theoretical predictions are generally consistent with the empirically measured robustness. They are more accurate for DDIM than the other two samplers. The observed deviation may be due to multiple reasons. First, the survival function (see red curve in Fig. 3b)  changes rapidly around CR = 0.2. Second, the CR of the image neighborhood is not uniform for all perturbations.

We conduct further analysis to generate predictions on the ImageNet dataset using DDPM and two additional continuous-time sampling methods (VPSDE from~\cite{song2020score}, and DPM from~\cite{lu2022dpm}). Across all diffusion models, substantial compression of the image space was observed (Table~\ref{tab:cr-imagenet-pred}). Based on these measured CR, our theory predicts that DDPM should exhibit slightly higher robustness than VPSDE, and both of them should substantially outperform DPM-20 in robustness. These provide testable predictions for future work.

\begin{table}[t]
\caption{Compression rates and robustness predictions on CIFAR-10.}
\label{tab:cr-cifar10-pred}
\centering
\small
\begin{tabular}{l|ccc}
\toprule[1.2pt]
\textbf{Metric} & \textbf{DDPM} & \textbf{Reverse-only} & \textbf{DDIM} \\
\midrule
Clean Acc. & 85.5\% & 87.2\% & 87.5\% \\
Expected CR & 0.231 & 0.195 & 0.281 \\
Expected CR (std) & 0.046 & 0.046 & 0.048 \\
\midrule
Pred. (Clean) & 27.8\% & 45.3\% & 10.2\% \\
Pred. (Anchor) & 26.0\% & 40.9\% & 10.6\% \\
PGD Robust. & 23.7\% & 34.9\% & 10.9\% \\
\bottomrule[1.2pt]
\end{tabular}
\end{table}

\begin{table}[t]
\caption{Compression rates and robustness predictions on ImageNet.}
\label{tab:cr-imagenet-pred}
\centering
\small
\begin{tabular}{l|ccc}
\toprule[1.2pt]
\textbf{Metric} & \textbf{DDPM} & \textbf{VPSDE} & \textbf{DPM-20} \\
\midrule
Clean Acc. & 73.6\% & 75.4\% & 59.8\% \\
Expected CR & 0.147 & 0.165 & 0.337 \\
Expected CR (std) & 0.070 & 0.059 & 0.015 \\
\midrule
Pred. (Clean) & 46.1\% & 40.1\% & 1.0\% \\
Pred. (Anchor) & 42.2\% & 38.0\% & 0.5\% \\
\bottomrule[1.2pt]
\end{tabular}
\end{table}

\section{Theoretical Analysis of the Compression Process}
\label{sec:theory}
\subsection{The compression process in a Gaussian score field}
We next perform theoretical analyses to understand the diffusion-model-induced compression effect reported above. To build intuition, we first examine the Gaussian score field. Assume the data $\bm{x}\in\mathbb{R}^d$ follow the multivariate Gaussian distribution, $\bm{x}\sim \mathcal{N}(\bm{\mu},\Sigma)$, where $\Sigma$ is the covariance matrix (symmetric and positive definite). The score function 
\begin{equation}
s(\bm{x})=\nabla_{\bm{x}}\log p(\bm{x})=-\Sigma^{-1}(\bm{x}-\bm{\mu}).
\end{equation}
Assume a dynamical system follows the score field, and discretize with the forward Euler method,
\begin{equation}
\dfrac{d\bm{x}}{dt}=s(\bm{x})\,\,\Rightarrow\,\,\bm{x}_{t+1}=f_t(\bm{x}_t)=\bm{x}_t-h_t\cdot\Sigma^{-1}(\bm{x}_t-\bm{\mu}),
\end{equation}
where $h_t$ is the step size at timestep $t$. Suppose the covariance matrix $\Sigma$ has eigenvector $\bm{v}_i$ with eigenvalue $\sigma^2_i$, the CR along the eigenvector, and the expected CR are given by (Appendix~A-B)
\begin{equation}
\textrm{CR}[f_t,\bm{\eta}\parallelsum \bm{v}_i]=1 - \frac{h_t}{\sigma_i^2},\quad \overline{\mathrm{CR}}[f_t]\approx 1 - \frac{h_t}{d}\sum_i \frac{1}{\sigma_i^2}.
\end{equation}
The results show a strong compression effect (small CR) along directions of low variance (off-manifold) and a weaker compression effect (large CR) along directions of high variance (on-manifold; see Fig.~\ref{fig:theory}a).

Natural images are often modeled under the low-dimensional manifold hypothesis, meaning that images lie in a low-dimensional subspace embedded in a high-dimensional space~\citep{simoncelli2001natural, maaten2008visualizing}. Consistent with this hypothesis, \citet{pope2021intrinsic} estimated the intrinsic dimension of CIFAR-100 to be around 30 and ImageNet to be around 45, both much lower than their corresponding embedding dimensions. This low-dimensional structure has also been central to recent theoretical analyses of diffusion models, including studies of score approximation and estimation on low-dimensional data~\citep{chen2023score}, the effectiveness of Gaussian score approximations~\citep{wang2024gaussian}, and the geometry-adaptive representations underlying their generalization~\citep{kadkhodaiegeneralization}. As a first step toward understanding how the low-dimensionality may affect compression, we study the case that the covariance matrix has an intrinsic dimension $d_{\textrm{in}}\ll d$, with large variances $\sigma_l^2$ along these intrinsic dimensions (i.e., on-manifold directions) and small variances $\sigma_s^2$ along the remaining ones (i.e., off-manifold directions). That is, the covariance matrix can be diagonalized as $
\Lambda=\textrm{diag}\big( \sigma_l^2 I_{d_{\mathrm{in}}},\;
\sigma_s^2 I_{d-d_{\mathrm{in}}} \big)
$, with $d_{\textrm{in}}\ll d$, $\sigma_l^2\gg\sigma_s^2$.
Under these assumptions, we find that the compression rates
\begin{equation}
\begin{aligned}
\textrm{CR}_{\textrm{on-manifold}}&=\textrm{CR}[f_t,\bm{\eta}\parallelsum\bm{v}_l]=1 - \frac{h_t}{\sigma^2_l},\\
\quad\textrm{CR}_{\textrm{off-manifold}}&=\textrm{CR}[f_t,\bm{\eta}\parallelsum\bm{v}_s]\approx \overline{\textrm{CR}}.
\end{aligned}
\end{equation}
The dominant compression effect along off-manifold directions is well captured by the expected CR. We illustrate this by empirically tracking the CR over time for a random initial perturbation as the system evolves. As shown in Fig.~\ref{fig:theory}b, the CR begins at the off-manifold value and eventually converges to the on-manifold value, demonstrating that off-manifold perturbations are compressed while on-manifold components are preserved, consistent with the accuracy of our approximation.

\subsection{Mathematical analysis of the compression process under diffusion models}
\begin{figure*}[ht]
\centering
\includegraphics[width=0.99\linewidth]{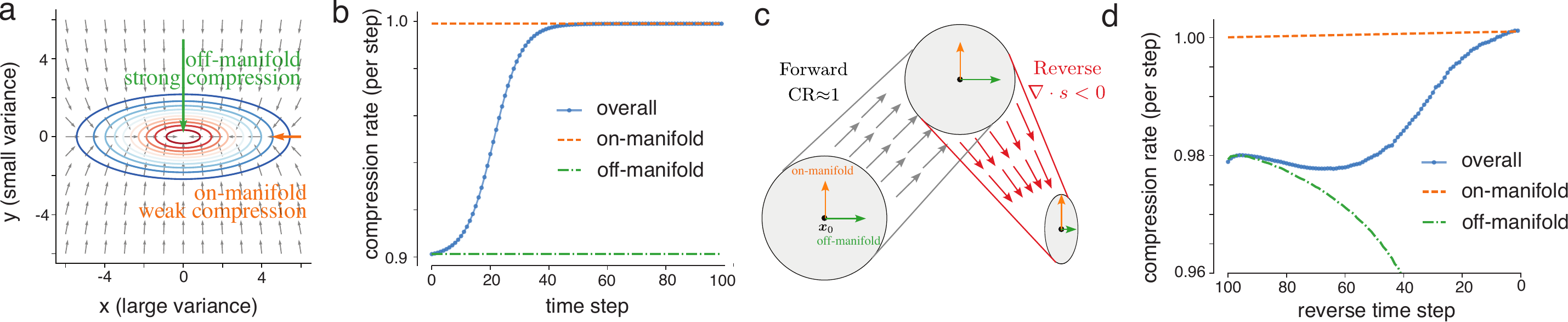}
\caption{\textbf{Theoretical analysis of the compression process.}
(a) Score field of a multivariate Gaussian distribution. The Gaussian exhibits constant CR along each direction, with small compression on-manifold and large compression off-manifold.
(b) As the system evolves through the Gaussian score field, CR transitions from the off-manifold value to the on-manifold value.
(c) Illustration of compression in diffusion models. The reverse process contributes most of the compression effect, primarily acting on off-manifold perturbations. (d) Measured CR at each step of diffusion models. Similar to the Gaussian case, CR transitions from the off-manifold to the on-manifold value. The initial match of the CR (blue) with the off-manifold value (green) supports our approximation in Eq.~\ref{eq:cr}. As the reverse process progresses, the two curves diverge, reflecting the perturbations being purified from off- to on-manifold directions.}
\label{fig:theory}
\end{figure*}

\paragraph{Compression induced by the convergent score field of reverse process.} We next extend our analyses to diffusion models. Calculate the CR of forward process at timestep $t$ with randomness $\bm{\epsilon}_t$
\begin{equation}
\textrm{CR}[f^{\textrm{FWD}}_{t,\bm{\epsilon}_t},\bm{x}_t,\bm{\eta}]=\sqrt{\alpha_t}\approx1.
\end{equation}
The forward process is primarily a translational shift, producing uniform compression across all directions (Fig.~\ref{fig:theory}c). The total compression induced by the forward process is bounded by $\bar{\alpha}_t=\prod_{t=1}^{t^*}\sqrt{\alpha_t}\approx 0.90$, which is not sufficient to yield substantial robustness gains according to our theory (the robustness gain is prominent only after reaching a CR of around 0.3, Fig.~\ref{fig:relation}b).

Let $\kappa_t=1/\sqrt{\alpha_t}$, $\gamma_t=(1-\alpha_t)/(\sqrt{1-
\bar{\alpha}_t})$, the expected CR of the reverse process at timestep $t$ is given by (Appendix~A-C),
\vspace{-1mm}
\begin{equation}
\label{eq:cr}
\overline{\textrm{CR}}[f^{\textrm{REV}}_{t,\bm{z}_t},\bm{x}_t]\approx\kappa_t\left(1-\frac{\gamma_t}{d}\nabla\cdot\boldsymbol{\epsilon}_{\theta}(\boldsymbol{x}_t,t)\right),
\end{equation}
where $\nabla\cdot\boldsymbol{\epsilon}_{\theta}(\boldsymbol{x}_t,t)=\tr[{J_{\boldsymbol{\epsilon}_\theta}(\boldsymbol{x}_t,t)}]$ is the divergence of the noise predictor. In practice, $\kappa_t$ is close to 1. This implies that a compression effect arises when the divergence $\nabla \cdot \boldsymbol{\epsilon}_\theta > 0$. By Tweedie’s formula~\citep{robbins1956empirical, miyasawa1961empirical}, the score function at $\boldsymbol{x}_t$ can be estimated using the optimal noise predictor (Appendix~A-A).
\vspace{-1mm}
\begin{equation}
s(\boldsymbol{x}_t)\approx-\frac{1}{\sqrt{1-\bar{\alpha}_t}}\boldsymbol{\epsilon}_\theta(\boldsymbol{x}_t,t).
\end{equation}
Thus, in the reverse process of diffusion models, the compression effect is induced by the convergent nature of the learned score function, $\nabla \cdot s(\bm{x}_t) < 0$.

\paragraph{Diffusion model purifies off-manifold while preserves on-manifold perturbations.}Analogous to the Gaussian case, define the eigendirection with small absolute eigenvalue $|\tilde{\lambda}_s|$ of the Jacobian of the score $J_s(\bm{x})$ as on-manifold, and the large absolute eigenvalue $|\tilde{\lambda}_l|$ direction as off-manifold.
Following the low-dimensional data manifold hypothesis, we assume that the eigenvalue matrix has the structure
\begin{equation}
\Lambda_{J_s(\bm{x})}=\textrm{diag}\big(\tilde{\lambda}_s I_{d_{\textrm{in}}},\tilde{\lambda}_lI_{d-d_{\textrm{in}}}\big),\quad d_{\textrm{in}}\ll d,\,\,|\tilde{\lambda}_s|\ll|\tilde{\lambda}_l|.
\end{equation}
We can approximate the CR along both on-manifold and off-manifold directions as (Appendix~A-C)
\begin{equation}
\textrm{CR}_{\textrm{on-manifold}}\approx\kappa_t,\,\,\,\,\textrm{CR}_{\textrm{off-manifold}}\approx\overline{\textrm{CR}}[f^{\textrm{REV}}_{t,\bm{z}_t},\bm{x}_t].
\end{equation}
We again empirically validate our theory by tracking the CR over time for a random initial perturbation under diffusion purification. As shown in Fig.~\ref{fig:theory}d, similar to the Gaussian case, during diffusion purification the CR initially reflects off-manifold value and gradually approaches the on-manifold value, indicating that off-manifold perturbations are selectively compressed while on-manifold components are preserved. The initial match of the CR (blue) with the off-manifold value (green) further supports the accuracy of our first-order approximation in Eq.~\ref{eq:cr}. As the reverse process progresses, the two curves diverge because the perturbations are transformed from off- to on-manifold directions.

\section{Discussions}

We have systematically studied how diffusion models improve adversarial robustness. Our end-to-end analysis leads to new insights into the mechanisms underlying the robustness improvement under diffusion-based purification by partitioning it into two sources. We find that diffusion models push perturbed images away from clean samples, rejecting major previous hypotheses on this question. We further find that a hyperspherical cap model approximately captures the shape of the adversarial regions for the base classifier. Thus, the adversarial regions are relatively large continuous regions, rather than highly scattered and small droplets as previously hypothesized~\cite{szegedy2013intriguing}. This observation implies that adversarial regions are difficult to escape by adding small Gaussian noise. It further predicts that the cosine similarity of actual vs. optimal attack under fixed-randomness should be predictive of the robustness improvement due to randomness (that can not be removed by EOT) in diffusion models. This prediction was confirmed in our empirical result. 

While our empirical results reported in Sec.~\ref{sec:random-residual} are well consistent with a hyperspherical cap model for the adversarial regions, Fig.~\ref{fig:hypercap}c suggests slight deviations from linear decision boundaries. We further investigate the implication of the deviation from linearity (Appendix~D-E), and find that adversarial attacks indeed exploit the nonlinear decision boundary. However, the systematic decrease in randomness residual with increased cosine similarity with PGD-Fix still holds.

Our analysis reveals a connection between the compression rate and the divergence of the score fields of diffusion models. As diffusion models trained on natural images generally show a compression rate small than 1 empirically, it suggests that the score fields are generally convergent in most areas of the image space (although unlikely to be true for all images). Our results on how compression rate changes as the reverse process unfolds show systematic deviations from the prediction of Gaussian score fields. This implies that the local properties of image distribution in the vicinity of image manifold may not be well approximated by Gaussian distributions. It would be interesting for future research to investigate which distributions better capture the dynamics of compression rate.

Importantly, when the randomness in diffusion models is fixed, the compression rate and empirical robustness follow a systematic relationship, providing strong support for our hypothesis that compression of the image space underlies the robustness improvements offered by diffusion models. Our results suggest a promising direction for designing “compression-based purification” systems, which should satisfy two criteria: (i) high clean accuracy at anchor points, and (ii) strong compression rates around anchor points. Systems meeting these criteria are expected to achieve reliable adversarial robustness without relying on stochasticity. 

In this study, we focus on diffusion models constructed in the image space, as most prior studies on diffusion-based purification have used this type of model~\citep{nie2022diffusion,xiao2023dense,lee2023robust,liu2025towards}. Recently, latent diffusion models~\citep{Rombach2022high} and flow-matching models~\citep{lipman2023flow} have also been shown to improve adversarial robustness~\citep{zhangclipure,collaert2025flowpure}. It would be interesting for future work to test whether our findings would generalize to these alternative purification models.

As the reverse process of diffusion models approximately implements Bayesian inference~\cite{xiao2023dense} under certain conditions, one interpretation of the robustness improvement under diffusion-based purification is that natural image prior suppresses nuisance variability to increase robustness. An influential theory of how the human visual systems processes visual inputs is that it performs hierarchical Bayesian inference~\cite{Lee2003hierarchical}. Substantial empirical evidence suggests that human performs Bayesian inference in vision tasks~\cite{Knill1996perception,Weiss2002motion,Kersten2004object,Hahn2024unifying}, although the exact implementation of Bayesian inference in neural circuits remain unclear. Given that our visual systems remain substantially more robust than the current AI systems, it is interesting to compare diffusion-based purification with hierarchical Bayesian theory of visual processing. 
To this end, a diffusion model trained in pixel space is akin to a brain that only performs Bayesian inference at the level of the retina (the first stage of visual processing), while a latent diffusion model is analogous to a brain that only performs Bayesian inference in some deeper layers. Based on this comparison, one promising path towards constructing more robust vision model might be to construct models that perform diffusion-based Bayesian inference at every stage of the processing, similar to what visual systems in the brain are hypothesized to be.

\section*{Acknowledgments}
We thank Nikolaus Kriegeskorte and Kevin Tian for valuable discussions and feedback.

Large language models (LLMs) were not used in the preparation of the initial manuscript or codebase. During later stages of revision, LLMs (ChatGPT~\citep{openai2023chatgpt} and Claude~\citep{anthropic2024claude3}) were used primarily for language editing and codebase organization. AI-assisted content generation was limited to Sec.~\ref{sec:random-residual}, and all generated content was reviewed and verified by the authors.

The authors declare no conflict of interest.

\renewcommand\bibsection{\section*{References}}
\putbib[diffusion-robust]
\end{bibunit}

\section{Biography Section}

\vspace{11pt}

\vspace{-33pt}
\begin{IEEEbiographynophoto}{Liu Yuezhang}
is currently a Ph.D. candidate in Neuroscience at The University of Texas at Austin. He received his B.Eng. in Automation from Tsinghua University in 2019 and his M.Phil. in Computer Science from The Hong Kong University of Science and Technology in 2021. His research lies in computational neuroscience, focusing on the dynamics and geometry of robust computation in perceptual systems. He is a recipient of the Joseph Needham Merit Scholarship.
\end{IEEEbiographynophoto}

\begin{IEEEbiographynophoto}{Xue-Xin Wei}
is an Assistant Professor at The University of Texas at Austin. He got his Bachelor's degree in Mathematics and Applied Mathematics from Peking University and his PhD in Psychology at University of Pennsylvania, respectively. His research seeks to understand how biological and artificial intelligence systems solve tasks efficiently, robustly and flexibly. He is the recipient of the Louis Flexner thesis Award in 2015, and a recipient of the Sloan Research Fellowship in 2025.
\end{IEEEbiographynophoto}

\vfill

\clearpage
\onecolumn

\makeatletter
\global\let\addcontentsline\oldaddcontentsline
\makeatother
\setcounter{tocdepth}{2}
\makeatletter
\renewcommand{\l@paragraph}[2]{}
\renewcommand{\l@subparagraph}[2]{}
\makeatother

\begingroup
\centering
\Huge\selectfont
Supplementary Material for\\``Demystifying Adversarial Robustness in Diffusion Models: Compression, Randomness, and Geometry''\par
\vspace{8.5pt}
\sublargesize\selectfont Liu~Yuezhang~and~Xue-Xin~Wei\par
\endgroup

\vspace{47pt}
{\hypersetup{linkcolor=black}%
\tableofcontents}

\clearpage
\begin{bibunit}
\appendices
\newcommand{\beginsupplement}{%
\setcounter{table}{0}
\renewcommand{\thetable}{S\arabic{table}}%
\setcounter{figure}{0}
\renewcommand{\thefigure}{S\arabic{figure}}%
}
\beginsupplement

\section{Theoretical results}
\subsection{Score function of DDPM via Tweedie's formula}
\label{app:tweedie}
\begin{proof}
The score function of the noisy sample $\bm{x}_t$ is defined as 
\[
s(\bm{x}_t) = \nabla_{\bm{x}_t} \log p(\bm{x}_t).
\]

By Tweedie's formula~\citep{robbins1956empirical, miyasawa1961empirical}, for a Gaussian observation 
\[
\bm{x}_t \sim \mathcal{N}(\bm{\mu}, \sigma^2 \bm{I}),
\]
the posterior mean $\mathbb{E}[\bm{\mu} \mid \bm{x}_t]$ can be expressed in terms of the score function:
\[
\mathbb{E}[\bm{\mu} \mid \bm{x}_t] = \bm{x}_t + \sigma^2 \nabla_{\bm{x}_t} \log p(\bm{x}_t).
\]

In the DDPM forward process (eq. 3), we can rewrite $\bm{x}_t$ as
\[
\bm{x}_t = \sqrt{\bar{\alpha}_t} \bm{x}_0 + \sqrt{1-\bar{\alpha}_t}\,\bm{\epsilon}, \quad \bm{\epsilon} \sim \mathcal{N}(\bm{0},\bm{I}),
\]
where $\sqrt{\bar{\alpha}_t}\bm{x}_0$ plays the role of the unknown mean $\bm{\mu}$ and $\sqrt{1-\bar{\alpha}_t}$ is the standard deviation.

Applying Tweedie's formula gives
\[
\nabla_{\bm{x}_t} \log p(\bm{x}_t) = \frac{\mathbb{E}[\sqrt{\bar{\alpha}_t}\,\bm{x}_0 \mid \bm{x}_t]-\bm{x}_t}{1-\bar{\alpha}_t}.
\]

The DDPM reverse process (eq. 4) estimates $\bm{x}_0$ using the noise predictor $\bm{\epsilon}_\theta(\bm{x}_t,t)$:
\[
\sqrt{\bar{\alpha}_t}\,\bm{x}_0 \approx \bm{x}_t - \sqrt{1-\bar{\alpha}_t}\,\bm{\epsilon}_\theta(\bm{x}_t,t).
\]

Plugging this into the score function expression yields
\[
\begin{aligned}
s(\bm{x}_t) &= \nabla_{\bm{x}_t} \log p(\bm{x}_t) \\
&\approx \frac{\bm{x}_t - \sqrt{1-\bar{\alpha}_t}\,\bm{\epsilon}_\theta(\bm{x}_t,t) - \bm{x}_t}{1-\bar{\alpha}_t} \\
&= - \frac{1}{\sqrt{1-\bar{\alpha}_t}} \, \bm{\epsilon}_\theta(\bm{x}_t,t).
\end{aligned}
\]

Hence, the DDPM noise predictor $\bm{\epsilon}_\theta(\bm{x}_t,t)$ provides a direct estimate of the score function up to a scaling factor.
\end{proof}

\subsection{Compression in the Gaussian score field}
\label{app:score-gaussian}
\begin{proof}
Let $\bm{x}\sim \mathcal{N}(\bm{x};\boldsymbol{\mu},\Sigma)$,
\[
p(\bm{x}) = \frac{1}{(2\pi)^{d/2}|\Sigma|^{1/2}}
\exp\!\Big(-\tfrac{1}{2}(\bm{x}-\boldsymbol{\mu})^T \Sigma^{-1}(\bm{x}-\boldsymbol{\mu})\Big),
\]
where \(\Sigma\) is a positive-definite covariance matrix.

The score function
\[
s(\bm{x}) \;=\; \nabla_{\bm{x}}\log p(\bm{x})
\;=\; -\frac{1}{2}\nabla_{\bm{x}\,}\big[(\bm{x}-\boldsymbol{\mu})^T \Sigma^{-1}(\bm{x}-\boldsymbol{\mu})\big]
\;=\; -\Sigma^{-1}(\bm{x}-\boldsymbol{\mu}).
\]

The Jacobian of the score
\[
J_s(\bm{x})
= \dfrac{\partial}{\partial \bm{x}}\big[-\Sigma^{-1}(\bm{x}-\boldsymbol{\mu})\big]
= -\Sigma^{-1}.
\]
which is constant and negative definite.

Since the covariance matrix is symmetric positive definite, diagonalize $\Sigma=V\Lambda V^T$,
\[
\Lambda=\textrm{diag}(\sigma_1^2,\dots,\sigma_d^2),\,\,V=(\bm{v}_1,\dots,\bm{v}_d),
\]
where $V$ is the orthonormal eigenvector matrix. The inverse of the covariance $\Sigma^{-1}$ can be diagonalized with the same eigenvectors as 
\[
\Sigma^{-1}=V\Lambda^{-1}V^T,\,\, \Lambda^{-1}=\textrm{diag}\left(\frac{1}{\sigma^2_1},\dots,\frac{1}{\sigma^2_d}\right).
\]
Therefore, the compression rate along a certain eigenvector $\bm{v}_i$ of $\Sigma$ is
\[
\textrm{CR}[f_t,\bm{\eta}\parallelsum \bm{v}_i]=\frac{\|(I_d-h_t\Sigma^{-1})\bm{\eta}\|}{\|\bm{\eta}\|}=1 - \frac{h_t}{\sigma_i^2}.
\]
For a perturbation $\bm{\eta}$ in general, decomposite onto the eigenvectors, $\bm{\eta}=\sum_i a_i\bm{v}_i$. Therefore the noise residual (numerator)
\[
f_t(\bm{x}_0+\bm{\eta})-f_t(\bm{x}_0)=(I_d-h_t\Sigma^{-1})\bm{\eta}=\sum_i \left(1-\frac{h_t}{\sigma_i^2}\right)a_i\bm{v}_i.
\]
For isotropic noise, $a_i=1/d$, $\|{\bm{\eta}}\|=\sqrt{1/d}$ is almost constant for large $d$, therefore the expected CR
\[
\overline{\textrm{CR}}=\sqrt{\sum_i \left(1-\frac{h_t}{\sigma_i^2}\right)^2\frac{1}{d}}=\sqrt{1-2\frac{h_t}{d}\sum_i\frac{1}{\sigma_i^2}+\frac{h_t^2}{d}\sum_i\frac{1}{\sigma_i^4}}.
\]
Since the timestep $h_t\to0$, $d\to\infty$, $\sqrt{1+u}\approx1+u/2$ ($u\to 0$), consequently
\[
\overline{\textrm{CR}}\approx 1-\frac{h_t}{d}\sum_i\frac{1}{\sigma_i^2}.
\]
Define the direction of large variance $\sigma_L^2$ as the on-manifold direction, and the small variance $\sigma_s^2$ direction as off-manifold. By the low-dimensional manifold assumption, the eigenvalue matrix has the structure
\[
\Lambda=\textrm{diag}\big(\underbrace{\sigma_l^2,\dots,\sigma_l^2}_{d_{\textrm{in}}},\underbrace{\sigma_s^2,\dots,\sigma_s^2}_{d-d_{\textrm{in}}}\big),
\]
with $d_{\textrm{in}}\ll d$, $\sigma_l^2\gg\sigma_s^2$. Therefore,
\[
\begin{aligned}
\textrm{CR}_{\textrm{on-manifold}}&=\textrm{CR}[f_t,\bm{\eta}\parallelsum\bm{v}_l]=1 - \frac{h_t}{\sigma^2_l},\\
\textrm{CR}_{\textrm{off-manifold}}&=\textrm{CR}[f_t,\bm{\eta}\parallelsum\bm{v}_s]=1 - \frac{h_t}{\sigma^2_s}\approx 1-\frac{d-d_\textrm{in}}{d}\frac{h_t}{\sigma_s^2}\approx\overline{\textrm{CR}}.
\end{aligned}
\]
\end{proof}
The hyperparameters used to simulate the Gaussian score field in Fig.~6 are listed in Table.~\ref{tab:hyperparameter-gaussian}.

\begin{table}[htbp]
\caption{Hyperparameters for simulating the Gaussian score field.}
\label{tab:hyperparameter-gaussian}
\centering
\begin{tabular}{l|c}
\toprule[1.5pt]
{\textbf{Hyperparameters}} & \textbf{Values} \\
\midrule
Dimension $d$  & 3072 \\
Intrinsic dimension $d_{\textrm{in}}$      & 40 \\
On-manifold std $\sigma_l$ & 1.0 \\
Off-manifold std $\sigma_s$ & 0.01 \\
Stepsize $h_t$  & 0.001 \\
Timesteps $T$   & 100 \\
\bottomrule[1.5pt]
\end{tabular}
\end{table}

\subsection{Compression in diffusion models}
\label{app:cr}
\begin{proof}
DiffPure~\citep{nie2022diffusion} runs the forward and reverse process of the diffusion model to an intermediate $t^*$ step as a purification system,
\[
f^{\textrm{DiffPure}}(\bm{x})=f_1^{\textrm{REV}}\circ\dots\circ f_{t^*}^{\textrm{REV}}\circ f_{t^*}^{\textrm{FWD}}\dots\circ f_1^{\textrm{FWD}}(\bm{x}).
\]
Following the definition, calculate the CR of forward process (eq.~3) at timestep $t$ with randomness $\bm{\epsilon}_t$
\[
\textrm{CR}[f^{\textrm{FWD}}_{t,\bm{\epsilon}_t},\bm{x}_t,\bm{\eta}]=\frac{\|\alpha_t\bm{\eta}\|}{\|\bm{\eta}\|}=\alpha_t\approx1.
\]
Thus, the forward process is primarily a translational shift, producing slight uniform compression across all directions.

Denote the reverse step of the diffusion model at timestep $t$ (eq.~4) as
\[
\boldsymbol{x}_{t-1}=f_t(\bm{x}_t)=\kappa_t\left(\boldsymbol{x}_t-\gamma_t\boldsymbol{\epsilon}_\theta(\boldsymbol{x}_t, t)\right)+\sigma_t\boldsymbol{z}_t,
\]
where $\kappa_t=1/\sqrt{\alpha_t}$, $\gamma_t=(1-\alpha_t)/(\sqrt{1-
\bar{\alpha}_t})$. Consider the Jacobian of the transformation $f_t$
\[
J_{f_t}(\bm{x}_t)=\kappa_t(I_d-\gamma_t J_{\bm{\epsilon}_\theta}(\bm{x}_t,t)).
\]
Suppose the Jacobian of the noise predictor $J_{\boldsymbol{\epsilon}_\theta}(\boldsymbol{x}_t,t)$ has eigenvector $\bm{v}_i$ with eigenvalue $\lambda_i$, then
\[
J_{f_t}(\bm{x}_t)\bm{v}_i=\kappa_t(1-\gamma_t\lambda_i)\bm{v}_i.
\]
At timestep $t$, with fixed randomness $\bm{z}_t$, the noise residual (numerator) with infinitesimal perturbation $\epsilon\to0$,
\[
f_t(\bm{x}_t+\epsilon\bm{\eta})-f_t(\bm{x}_t)\approx J_{f_t}(\bm{x}_t)\epsilon\bm{\eta}.
\]
Therefore the CR along the eigendirection $\bm{v}_i$,
\[
\textrm{CR}[f_{t,\bm{z}_t}^{\textrm{REV}},\bm{x}_t,\epsilon\bm{v}_i]\approx\kappa_t(1-\gamma_t\lambda_i).
\]
For isotropic noise, decomposite $\bm{\eta}$ onto eigendirections, $\bm{\eta}=\sum_i a_i\bm{v}_i$, where $a_i=1/d$, $\|{\bm{\eta}}\|=\sqrt{1/d}$ is almost constant for large $d$. Therefore the expected CR at timestep $t$
\[
\begin{aligned}
\overline{\textrm{CR}}[f^{\textrm{REV}}_{t,\bm{z}_t},\bm{x}_t]
&\approx \kappa_t\sqrt{\frac{1}{d}\sum_i(1-\gamma_t\lambda_i)^2} \\
&=\kappa_t\sqrt{
1 - \frac{2\gamma_t}{d}\left(\sum_i\lambda_i\right)
+ \frac{\gamma_t^2}{d}\left(\sum_i\lambda_i^2\right)} \\
&=
\kappa_t\sqrt{
1 - \frac{2\gamma_t}{d}\,\nabla\cdot\boldsymbol{\epsilon}_{\theta}(\boldsymbol{x}_t,t)
+ \frac{\gamma_t^2}{d}\|J_{\boldsymbol{\epsilon}_\theta}(\boldsymbol{x}_t,t)\|_F^2}.
\end{aligned}
\]
Since $\gamma_t\to0$, $d\to\infty$, $\overline{\textrm{CR}}$ can be further approximated as
\[
\overline{\textrm{CR}}\approx\kappa_t\left(1-\frac{\gamma_t}{d}\nabla\cdot\boldsymbol{\epsilon}_{\theta}(\boldsymbol{x}_t,t)\right).
\]

By Tweedie's formula, $\bm{v}_i$ is also an eigenvector of the Jacobian of the score function $J_s(\bm{x}_t)$, with the eigenvalue
\[
\tilde{\lambda}_i=-\frac{\lambda_i}{\sqrt{1-\bar{\alpha}_t}},\quad\tilde{\bm{v}}_i=\bm{v}_i.
\]

Define the eigendirection with small absolute eigenvalue $|\tilde{\lambda}_s|$ of the Jacobian of the score $J_s(\bm{x})$ as the on-manifold direction, and the large absolute eigenvalue $|\tilde{\lambda}_l|$ direction as off-manifold. By the low-dimensional manifold assumption, the eigenvalue matrix has the structure
\[
\Lambda_s(\bm{x})=\textrm{diag}\big(\underbrace{\tilde{\lambda}_s,\dots,\tilde{\lambda}_s}_{d_{\textrm{in}}},\underbrace{\tilde{\lambda}_l,\dots,\tilde{\lambda}_l}_{d-d_{\textrm{in}}}\big),
\]
with $d_{\textrm{in}}\ll d$, $|\tilde{\lambda}_s|\ll|\tilde{\lambda}_l|\Rightarrow |\lambda_s|\ll|\lambda_l|$. Consequently,
\[
\begin{aligned}
\textrm{CR}_{\textrm{on-manifold}}&=\textrm{CR}[f_t,\bm{\eta}\parallelsum\bm{v}_s]\approx\kappa_t(1-\gamma_t\lambda_s)\approx\kappa_t,\\
\textrm{CR}_{\textrm{off-manifold}}&=\textrm{CR}[f_t,\bm{\eta}\parallelsum\bm{v}_l]\approx\kappa_t(1-\gamma_t\lambda_l)\approx \kappa_t\left(1-\gamma_t\frac{d-d_\textrm{in}}{d}\lambda_l\right)\approx\overline{\textrm{CR}}.
\end{aligned}
\]
\end{proof}

\subsection{Variance decomposition in diffusion models for adversarial purification}
\label{app:var-decomp}
Assuming that inputs $\bm{x}$ are independent of randomness ${\boldsymbol{\xi}}$, the total variance of the diffusion models can then be decomposed as
\begin{equation}
\begin{aligned}
\mathrm{Var}_{\boldsymbol{x},{\boldsymbol{\xi}}}[f(\boldsymbol{x},{\boldsymbol{\xi}})] = \mathbb{E}_{{\boldsymbol{\xi}}} \left[ \mathrm{Var}_{\boldsymbol{x}}f(\boldsymbol{x}|{\boldsymbol{\xi}}) \right]
+ \mathrm{Var}_{{\boldsymbol{\xi}}} \left[ \mathbb{E}_{\boldsymbol{x}}f(\boldsymbol{x}|{\boldsymbol{\xi}}) \right].
\end{aligned}
\label{eq:var-decomp1}
\end{equation}
The result directly follows the law of total variance. Here we append the proof for completeness.

\begin{proof}
Define the following means
\begin{align*}
\mu_{\boldsymbol{\xi}} &:= \mathbb{E}_{\boldsymbol{x}}f(\boldsymbol{x}|{\boldsymbol{\xi}}) \quad \text{(mean at fixed randomness } {\boldsymbol{\xi}}) \\
\mu &:= \mathbb{E}_{\boldsymbol{x}, {\boldsymbol{\xi}}}f(\boldsymbol{x}, {\boldsymbol{\xi}}) = \mathbb{E}_{\boldsymbol{\xi}}\mu_{\boldsymbol{\xi}} \quad \text{(global mean)}.
\end{align*}

Expand the total variance
\begin{align*}
\mathrm{Var}_{\boldsymbol{x}, {\boldsymbol{\xi}}}[f(\boldsymbol{x}, {\boldsymbol{\xi}})] = \mathbb{E}_{\boldsymbol{x}, {\boldsymbol{\xi}}} \left[ \| f(\boldsymbol{x}, {\boldsymbol{\xi}}) - \mu \|^2 \right] = \mathbb{E}_{\boldsymbol{\xi}} \left[ \mathbb{E}_{\boldsymbol{x}} \left[ \| f(\boldsymbol{x}, {\boldsymbol{\xi}}) - \mu \|^2 \right] \right].
\end{align*}

Now insert and subtract \( \mu_{\boldsymbol{\xi}} \) inside the norm:
\begin{align*}
\| f(\boldsymbol{x}, {\boldsymbol{\xi}}) - \mu \|^2 
&= \| f(\boldsymbol{x}, {\boldsymbol{\xi}}) - \mu_{\boldsymbol{\xi}} + \mu_{\boldsymbol{\xi}} - \mu \|^2 \\
&= \| f(\boldsymbol{x}, {\boldsymbol{\xi}}) - \mu_{\boldsymbol{\xi}} \|^2 
+ 2 \langle f(\boldsymbol{x}, {\boldsymbol{\xi}}) - \mu_{\boldsymbol{\xi}}, \mu_{\boldsymbol{\xi}} - \mu \rangle 
+ \| \mu_{\boldsymbol{\xi}} - \mu \|^2.
\end{align*}

Taking expectation over \( \boldsymbol{x} \) (for fixed \( {\boldsymbol{\xi}} \)), the cross term vanishes:
\[
\mathbb{E}_{\boldsymbol{x}}[ f(\boldsymbol{x}, {\boldsymbol{\xi}}) - \mu_{\boldsymbol{\xi}} ] = 0
\quad \Rightarrow \quad 
\mathbb{E}_{\boldsymbol{x}} \left[ \langle f(\boldsymbol{x}, {\boldsymbol{\xi}}) - \mu_{\boldsymbol{\xi}}, \mu_{\boldsymbol{\xi}} - \mu \rangle \right] = 0
.
\]

Therefore
\begin{align*}
\mathbb{E}_{\boldsymbol{x}} \left[ \| f(\boldsymbol{x}, {\boldsymbol{\xi}}) - \mu \|^2 \right]
= \mathbb{E}_{\boldsymbol{x}} \left[ \| f(\boldsymbol{x}, {\boldsymbol{\xi}}) - \mu_{\boldsymbol{\xi}} \|^2 \right] + \| \mu_{\boldsymbol{\xi}} - \mu \|^2.
\end{align*}

Finally take expectation over \( {\boldsymbol{\xi}} \):
\begin{align*}
\mathrm{Var}_{\boldsymbol{x}, {\boldsymbol{\xi}}}[f(\boldsymbol{x}, {\boldsymbol{\xi}})]
&= \mathbb{E}_{\boldsymbol{\xi}} \left[ \mathbb{E}_{\boldsymbol{x}} \left[ \| f(\boldsymbol{x}, {\boldsymbol{\xi}}) - \mu_{\boldsymbol{\xi}} \|^2 \right] \right] 
+ \mathbb{E}_{\boldsymbol{\xi}} \left[ \| \mu_{\boldsymbol{\xi}} - \mu \|^2 \right] \\
&= \mathbb{E}_{\boldsymbol{\xi}} \left[ \mathrm{Var}_{\boldsymbol{x}}f(\boldsymbol{x} | {\boldsymbol{\xi}}) \right] 
+ \mathrm{Var}_{\boldsymbol{\xi}} \left[ \mathbb{E}_{\boldsymbol{x}}f(\boldsymbol{x}|{\boldsymbol{\xi}}) \right].
\end{align*}

This completes the proof of the decomposition. 
\end{proof}

Based on the decomposition, the first term $\mathbb{E}_{\boldsymbol{\xi}} \left[ \mathrm{Var}_{\boldsymbol{x}}f(\boldsymbol{x} | {\boldsymbol{\xi}}) \right]$ represents the input variability, and the second term $\mathrm{Var}_{\boldsymbol{\xi}} \left[ \mathbb{E}_{\boldsymbol{x}}f(\boldsymbol{x}|{\boldsymbol{\xi}}) \right]$ represents the internal variability. Treating the input perturbations as signals and internal randomness as noise, we can further define the SNR in eq.~8.

\section{The hyperspherical cap model of adversarial regions}
\label{app:proof-cap}
We give a formal version of the hyperspherical cap model used
in Sec.~IV-C. The construction is stated
for an arbitrary classifier $g$ and applies, in particular,
to (i) a base CNN classifier acting directly on
$\boldsymbol{x}_0$, in which case the adversarial direction
is $\boldsymbol{\eta}_{\textrm{adv}}$, and (ii) a
fixed-randomness diffusion-purification system
$g\circ f_{\boldsymbol{\xi}_{\textrm{test}}}$, in which case
the adversarial direction is
$\boldsymbol{\eta}(\boldsymbol{\xi}_{\textrm{test}})$.

\begin{definition}[Adversarial directions]
Let $(\bm{x}_0,y_0)$ be a data point--label pair, and consider
its robustness in an $\ell_2$ ball of radius $\epsilon$. Let
$g(\bm{x})$ be a classifier correctly classifying the point,
i.e.\ $\arg\max_i g_i(\bm{x}_0)=y_0$, where $g_i(\bm{x})$
denotes the $i$-th logit. For a direction $\bm{\eta}$, define
the range of the $i$-th logit along $\bm{\eta}$ as
\[
r_i(\bm{\eta},\epsilon) =
\sup_{\gamma_1\in[0,\epsilon)} g_i\!\left(\bm{x}_0+\gamma_1 \tfrac{\bm{\eta}}{\|\bm{\eta}\|_2}\right)
- \inf_{\gamma_2\in[0,\epsilon)} g_i\!\left(\bm{x}_0+\gamma_2 \tfrac{\bm{\eta}}{\|\bm{\eta}\|_2}\right).
\]
A direction $\bm{\eta}_{\textrm{adv}}$ is an
\emph{adversarial direction} if for some logit $i$ the change
of logit along that direction is considerably larger than the
expectation along a random direction $\bm{\eta}_{\textrm{rand}}$:
\[
\mathbb{E}_{\bm{\eta}_{\textrm{rand}}} r_i(\bm{\eta}_{\textrm{rand}},\epsilon)
= o\!\left(r_i(\bm{\eta}_{\textrm{adv}},\epsilon)\right).
\]
\end{definition}

\begin{assumption}[Sparsity]
\label{assumption:sparse}
Adversarial directions are sparse: a random direction is, with
high probability, not an adversarial direction. This is
consistent with the fact that classifiers trained with random
Gaussian noise do not yield meaningful adversarial robustness.
\end{assumption}

\begin{assumption}[Critical threshold]
\label{assumption:ct}
Locally, when moving along an adversarial direction starting
from the clean sample, the decision boundary is crossed
exactly once. The radius at this single crossing is the
\emph{critical threshold} $\gamma$.
\end{assumption}

\begin{model}[Hyperspherical cap model of adversarial regions]
For small $\epsilon$, consider a point $\bm{x}$ whose
projection onto an adversarial direction
$\bm{\eta}_{\textrm{adv}}$ satisfies
\begin{equation}
\langle \bm{x}-\bm{x}_0,\, \bm{\eta}_{\textrm{adv}}/\|\bm{\eta}_{\textrm{adv}}\|_2 \rangle
= \beta\epsilon = O(\epsilon),\quad \beta\in\mathbb{R}.
\end{equation}
Under Assumption~\ref{assumption:sparse}, the logits at
$\bm{x}$ are well approximated by the logits along the
projection,
\begin{equation}
g(\bm{x}) \approx g\!\left(\bm{x}_0 + \beta\epsilon\cdot \bm{\eta}_{\textrm{adv}}/\|\bm{\eta}_{\textrm{adv}}\|_2\right).
\end{equation}
Combined with Assumption~\ref{assumption:ct}, $\bm{x}$ is an
adversarial example iff the projection crosses the critical
threshold, i.e.\ $\beta>\gamma/\epsilon$. The set of all such
points forms a hyperspherical cap around the adversarial
direction.
\end{model}

\begin{proof}
For a point $\bm{x}$ near $\bm{x}_0$, decompose
$(\bm{x}-\bm{x}_0)$ into components parallel and orthogonal to
$\bm{\eta}_{\textrm{adv}}$:
\[
(\bm{x}-\bm{x}_0)
= (\bm{x}-\bm{x}_0)_{\parallelsum\bm{\eta}_{\textrm{adv}}}
+ (\bm{x}-\bm{x}_0)_{\perp\bm{\eta}_{\textrm{adv}}},
\]
where $(\bm{x}-\bm{x}_0)_{\parallelsum\bm{\eta}_{\textrm{adv}}}
= \beta\epsilon\cdot\bm{\eta}_{\textrm{adv}}/\|\bm{\eta}_{\textrm{adv}}\|_2$.
Since $\epsilon$ is small, a Taylor expansion of logit $i$ at
$\bm{x}_0+(\bm{x}-\bm{x}_0)_{\parallelsum\bm{\eta}_{\textrm{adv}}}$
gives
\[
\begin{aligned}
g_i(\bm{x})
&= g_i\!\left(\bm{x}_0+(\bm{x}-\bm{x}_0)_{\parallelsum\bm{\eta}_{\textrm{adv}}}+(\bm{x}-\bm{x}_0)_{\perp\bm{\eta}_{\textrm{adv}}}\right)\\
&\approx g_i\!\left(\bm{x}_0+(\bm{x}-\bm{x}_0)_{\parallelsum\bm{\eta}_{\textrm{adv}}}\right)
+ \nabla^{\!\top} g_i\!\left(\bm{x}_0+(\bm{x}-\bm{x}_0)_{\parallelsum\bm{\eta}_{\textrm{adv}}}\right)\cdot(\bm{x}-\bm{x}_0)_{\perp\bm{\eta}_{\textrm{adv}}}.
\end{aligned}
\]
Assuming the gradients vary slowly in this small neighborhood,
\[
\left|\nabla^{\!\top} g_i\!\left(\bm{x}_0+(\bm{x}-\bm{x}_0)_{\parallelsum\bm{\eta}_{\textrm{adv}}}\right)\cdot(\bm{x}-\bm{x}_0)_{\perp\bm{\eta}_{\textrm{adv}}}\right|
\approx
\left|\nabla^{\!\top} g_i(\bm{x}_0)\cdot(\bm{x}-\bm{x}_0)_{\perp\bm{\eta}_{\textrm{adv}}}\right|
\le r_i\!\left(\perp\bm{\eta}_{\textrm{adv}},\epsilon\right).
\]
Under Assumption~\ref{assumption:sparse}, with high probability
the orthogonal direction $\perp\bm{\eta}_{\textrm{adv}}$ is not
itself an adversarial direction, so
\[
\mathbb{E}_{\perp\bm{\eta}_{\textrm{adv}}}\, r_i\!\left(\perp\bm{\eta}_{\textrm{adv}},\epsilon\right)
= o\!\left(r_i(\bm{\eta}_{\textrm{adv}},\epsilon)\right),
\]
yielding
$g_i(\bm{x})\approx
g_i\!\left(\bm{x}_0+\beta\epsilon\cdot\bm{\eta}_{\textrm{adv}}/\|\bm{\eta}_{\textrm{adv}}\|_2\right)$.
Combined with Assumption~\ref{assumption:ct}, $\bm{x}$ is
adversarial iff $\beta>\gamma/\epsilon$, which describes a
hyperspherical cap with axis $\bm{\eta}_{\textrm{adv}}$.
\end{proof}

\subsection{The GAAS arrangement (maximum $k$-simplex in a hyperspherical cap)}
\label{app:proof-gaas}
Here we provide an alternative proof of the GAAS algorithm~\citep{tramer2017space} based on the hyperspherical cap model.
\begin{proof}
For a $(k-1)$ simplex embedded in the $k$ dimensional space with radius $\epsilon$, the coordinates are the permutations of
\[
\left(0,\dots,\epsilon,\dots,0\right).
\]
The center is given by $\left(\frac{\epsilon}{k},\dots,\frac{\epsilon}{k}\right)$. Therefore the height (distance to the origin) is
\[
h=\frac{\epsilon}{\sqrt{k}}.
\]
With the hyperspherical cap model, the height should exceed the critical threshold $\gamma$, denote $\alpha=\gamma / \epsilon$,
\[
h=\frac{\epsilon}{\sqrt{k}}\geq\gamma\Rightarrow k\leq\frac{1}{\alpha^2}.
\]
Therefore the maximum number of orthogonal directions that can fit into the adversarial regions is 
\[
k=\min\left(\left\lfloor\frac{1}{\alpha^2}\right\rfloor,\,d\right),
\]
where $d$ is the dimensionality of the data as a trivial bound.
\end{proof}


\section{Implementation details of adversarial attacks on diffusion models}
\label{app:detail-attack}
\paragraph{Datasets and base classifiers}The experiments were conducted on the CIFAR-10~\citep{krizhevsky2009learning} and ImageNet~\citep{deng2009imagenet} datasets. For CIFAR-10, we subsampled the first 1000 images from the test set. For ImageNet, we subsampled the first 200 images from the validation set. Standard preprocessing was applied to the datasets. We used the standard classifiers from the \texttt{RobustBench}~\citep{croce2020robustbench}\url{https://github.com/RobustBench/robustbench}. Namely, the WideResNet-28-10 model for CIFAR-10, and ResNet-50 model for ImageNet. The clean accuracy on our subsampled set for the classifier is 94.8\% on CIFAR-10 (vs. 94.78\% on the full set) and 74.5\% on ImageNet (vs. 76.52\% on the full set).

\paragraph{Diffusion models}We focused on discrete-time diffusion models in this paper to avoid the potential gradient masking induced by numerical solvers in continuous-time models~\citep{pmlr-v145-huang22a}. For CIFAR-10, we used the official checkpoint of DDPM (converted to PyTorch from Tensorflow~\url{https://github.com/pesser/pytorch_diffusion}) instead of Score-SDE. For ImageNet, we used the official checkpoint of $256\times256$ unconditional Guided diffusion~\citep{dhariwal2021diffusion}~\url{https://github.com/openai/guided-diffusion} as the purification system. The purification time steps were kept the same with ~\citet{nie2022diffusion}, namely $t^*=0.1$ (100 forward and 100 reverse steps) for CIFAR-10 and $t^*=0.15$ (150 forward and 150 reverse steps) for ImageNet.

The DiffPure~\citep{nie2022diffusion} framework proposed to utilize both the forward and reverse processes of diffusion models for adversarial purification. Since the forward process introduces a large amount of randomness, we explore whether it's possible to remove the forward process, thus only using the reverse process of diffusion models for adversarial purification. A similar reverse-only framework was proposed in DensePure~\citep{xiao2023dense}, but further equipped with a majority voting mechanism to study the certified robustness.


\paragraph{Fixing randomness in diffusion models}
We controlled the randomness within diffusion models by controlling the random seeds during both the forward and reverse processes. For the base seed $s$, $i$-th batch of data at the $t$ step of the forward/reverse process, we set the random seed
\begin{equation}
\texttt{seed}(s,i,t) =
\left\{
\begin{aligned}
&\texttt{hash}(s,i,2t),\quad\quad\,\,\,\,\text{if forward process} \\
&\texttt{hash}(s,i,2t+1),\quad\text{if reverse process} \\
\end{aligned}
\right.
\end{equation}
before sampling the Gaussian noise from eq.~3 or eq.~4. The multiplicative hashing function
\begin{equation}
\texttt{hash}(s,i,t) = (p_s\cdot s) \oplus (p_i\cdot i) \oplus (p_t\cdot t) \mod 2^{32}
\end{equation}
where $p_s,p_i,p_t$ are large numbers coprime with each other to avoid collision and $\oplus$ denotes bitwise XOR. This setting ensures that we have a different random seed for each batch of data and timesteps in the forward/reverse process, but will keep the randomness the same through the entire purification process if encountering the same data batch.

\begin{table}[htbp]
\caption{Hyperparameters for adversarial attacks.}
\label{tab:hyperparameter}
\centering
\begin{tabular}{l|c}
\toprule[1.5pt]
{\textbf{Hyperparameters}} & \textbf{Values \{CIFAR-10, ImageNet\}} \\
\midrule
Attack magnitude  & \{8, 4\} / 255 \\
PGD steps        & 40 \\
Relative PGD step size & 0.01 / 0.3 \\
EOT numbers      & 15 \\
Batch size   & 1 \\
\midrule
Random factor $p_s$        & 83492791 \\
Random factor $p_i$        & 73856093 \\
Random factor $p_t$        & 19349663 \\
\bottomrule[1.5pt]
\end{tabular}
\end{table}

\paragraph{Adversarial attacks}We conducted BPDA/BPDA-EOT and PGD/PGD-EOT attacks~\citep{athalye2018obfuscated} on CIFAR-10 with $\ell_\infty=8/255$, and BPDA/BPDA-EOT attacks on ImageNet with $\ell_\infty=4/255$. The PGD was conducted based on the \texttt{foolbox}~\citep{rauber2017foolboxnative}\url{https://github.com/bethgelab/foolbox}, and the BPDA wrapper was adapted from \texttt{advertorch}~\citep{ding2019advertorch}\url{https://github.com/BorealisAI/advertorch}. Full gradients were calculated for the PGD/PGD-EOT as \citep{lee2023robust} discovered that the approximations methods used in the original DiffPure~\citep{nie2022diffusion} incurred weaker attacks. The full gradient of PGD/PGD-EOT is the strongest attack for DiffPure methods according to~\citet{lee2023robust} experiments, and is very computationally expensive. We ran our CIFAR-10 attack experiments on a NVIDIA RTX 6000 GPU for 10 days. We were not able to conduct the full PGD attack on ImageNet in a reasonable time given our available resources. 
The key hyperparameters for our attacks are listed in Table~\ref{tab:hyperparameter}. All attacks were repeated three times (n = 3) to compute the standard deviation, except for the EOT experiments, for which we were unable to do so due to limited computational resources. We used the base seeds $s=0,1,2$ for all experiments. We use additive uniform noise $U[-\epsilon,\epsilon]$ to estimate the expected CR, as it allows convenient control of the $\ell_\infty$ norm. Strictly speaking, uniform noise is not isotropic; however, we find empirically that the difference is negligible when $\epsilon$ is small.

\paragraph{FID score}The FID score was calculated based on the \texttt{pytorch-fid} package~\citep{Seitzer2020FID}\url{https://github.com/mseitzer/411
pytorch-fid}. We compute the FID score using 768-dimensional pre-classifier features due to the limited number of PGD samples.

\section{Additional experimental results}
\subsection{$\ell_p$ distance measurements during DiffPure}
\label{app:add-measure}
\begin{figure*}[h]
\centering
\subfloat[$\ell_2$ distances, CIFAR-10.]{
\includegraphics[width=0.45\columnwidth]{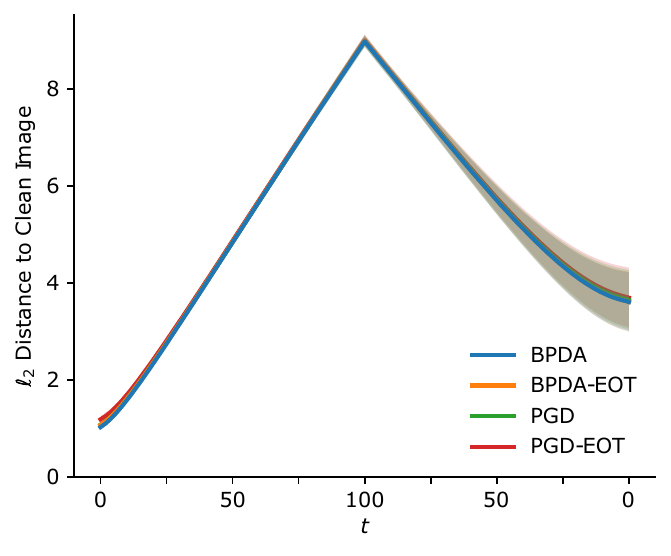}
\label{fig:l_cifar10_l2}
}
\hspace{2em}
\subfloat[$\ell_\infty$ distances, CIFAR-10.]{
\includegraphics[width=0.46\columnwidth]{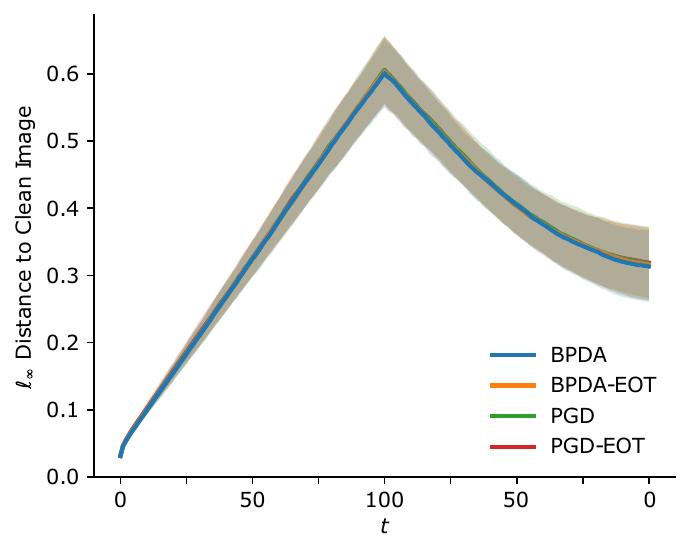}
\label{fig:l_cifar10_linf}
}
\caption{Additional distance measurements during DiffPure on CIFAR-10.}
\label{fig:l_cifar10}
\end{figure*}

\begin{figure*}[h]
\centering
\subfloat[$\ell_2$ distances, ImageNet.]{
\includegraphics[width=0.45\columnwidth]{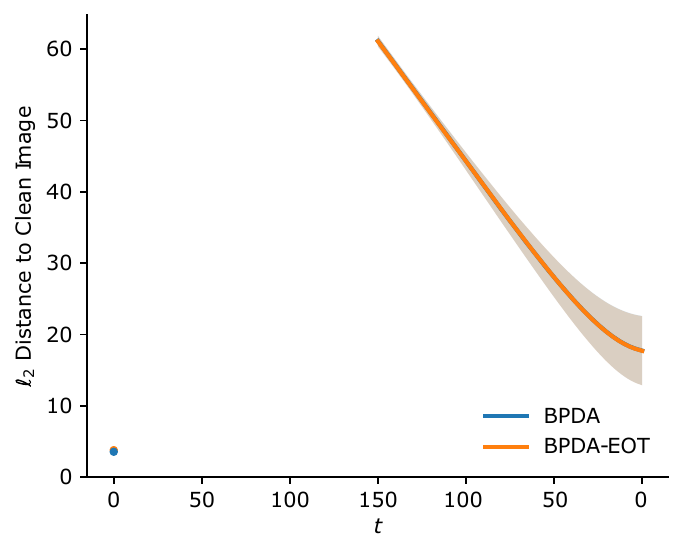}
\label{fig:l_imagenet_l2}
}
\hspace{2em}
\subfloat[$\ell_\infty$ distances, ImageNet.]{
\includegraphics[width=0.46\columnwidth]{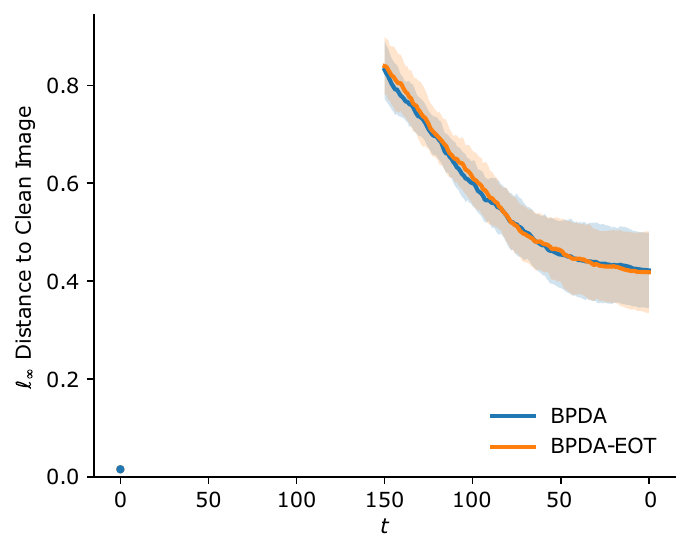}
\label{fig:l_imagenet_linf}
}
\caption{Additional distance measurements during DiffPure on ImageNet.}
\label{fig:l_imagenet}
\end{figure*}

Additional distance measurements during the DiffPure process are shown in Fig.~\ref{fig:l_cifar10} and~\ref{fig:l_imagenet}. For CIFAR-10, we further measured the $\ell_\infty$ distances for the experiment illustrated in Fig.~1c. For ImageNet, we repeated the same experiment with the unconditional Guided diffusion with 150 diffusion and denoising steps ($t^*=0.15$, the same setting with the DiffPure~\citep{nie2022diffusion}), and measured the $\ell_2/\ell_\infty$ distances. The distances during the intermediate diffusion process in ImageNet (Fig.~\ref{fig:l_imagenet}) are not shown as the code base implemented the one-step diffusion equation equivalent to the multistep diffusion. Again, similar effects were observed under both $\ell_2/\ell_\infty$ distances across datasets, namely, diffusion models purified to states further away from the clean images, considerably larger than the original adversarial perturbation ball. Detailed data points are listed in Table~\ref{tab:res-cifar10-l2},\ref{tab:res-cifar10-linf},\ref{tab:dist-imagenet}. Specifically, the $\ell_2/\ell_\infty$ distances to clean samples at the init point ($t=0$, the scale of the original perturbation), maximum point ($t=100/150$, after forward diffusion), and end point ($t=200/300$, after the reverse denoising). The end point distances are roughly 4 or 5 times of the size of the adversarial ball under $\ell_2$ distance on CIFAR-10/ImageNet, and 10 or 26 times under $\ell_\infty$ distance. Diffusion models transit back to the $\ell_2$ shrinkage regime beyond the uniform noise of $\epsilon=16/255$, which is twice of the standard $\ell_\infty$ adversarial ball considered for CIFAR-10.

%

\begin{table}[htbp]
\caption{$\ell_2/\ell_\infty$ distances measurements during DiffPure on ImageNet ($\ell_\infty=4/255$). 
}
\label{tab:dist-imagenet}
\centering
\begin{tabular}{cl|ccc}
\toprule[1.5pt]
{\textbf{Distances}} & {\textbf{Attack}} & \textbf{Init (\bm{$t=0$})} & \textbf{Max (\bm{$t=150$})} & \textbf{End (\bm{$t=300$})} \\
\midrule
\multirow{2}{*}{$\ell_2$} & {BPDA}       & 3.537 $\pm$ 0.079 & 61.116 $\pm$ 0.738 & 17.712 $\pm$ 4.851 \\
& {BPDA-EOT}   & 3.772 $\pm$ 0.139 & 61.078 $\pm$ 0.762 & 17.694 $\pm$ 4.838 \\
\midrule
\multirow{2}{*}{$\ell_\infty$} & {BPDA} & 0.016 $\pm$ 0.000 & 0.832 $\pm$ 0.059 & 0.422 $\pm$ 0.077 \\
& {BPDA-EOT}    & 0.016 $\pm$ 0.000 & 0.839 $\pm$ 0.060 & 0.418 $\pm$ 0.084 \\
\bottomrule[1.5pt]
\end{tabular}
\end{table}

\subsection{Behavior of diffusion models under random perturbations}
\label{app:uniform}
\begin{figure*}[h]
\centering
\subfloat[$\ell_2$ distances, uniform noise.]{
\includegraphics[width=0.45\columnwidth]{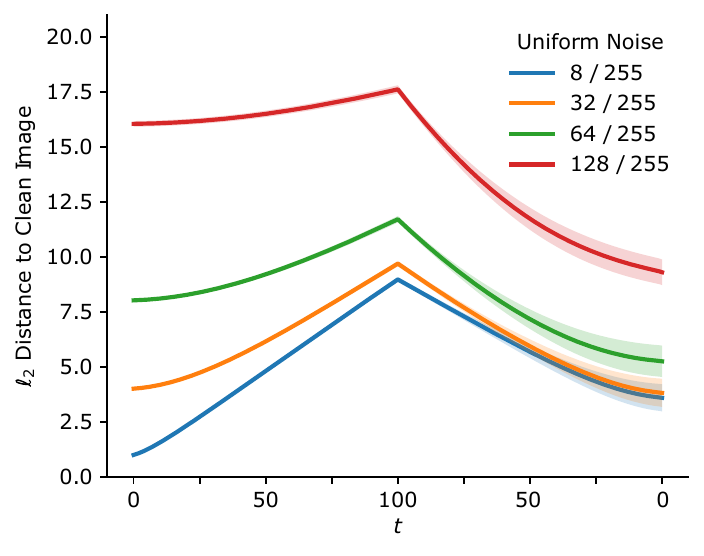}
\label{fig:l2_uniform}
}
\hspace{2em}
\subfloat[$\ell_\infty$ distances, uniform noise.]{
\includegraphics[width=0.45\columnwidth]{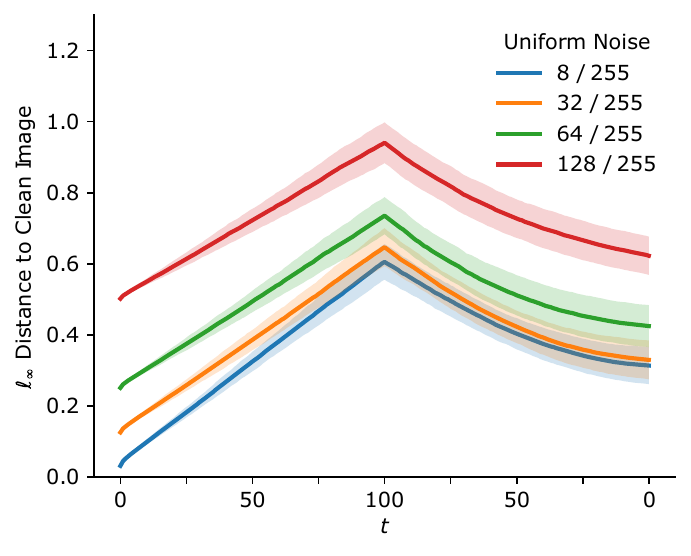}
\label{fig:linf_uniform}
}
\caption{Distance measurements under random perturbations on CIFAR-10.}
\label{fig:l_uniform}
\end{figure*}

We wonder if the behavior of adversarial attacks under diffusion models is special at all, that is, whether the push-away phenomena we observed are in fact general to arbitrary perturbations around the clean images. To test this, we generated perturbations of clean images by sampling random noise uniformly with a fixed magnitude. We first tested small perturbations that match the size of the adversarial attack on CIFAR-10 ($\ell_\infty=8/255$ uniform noise). We found that the behavior of the model under random noise (Fig.~\ref{fig:l_uniform}a, blue curve) is almost identical to that induced by adversarial attack (Fig.~1c, blue curve). These results, together with those reported above, suggest that diffusion models are not able to reduce the distances to a clean image from a slightly perturbed clean image. This raised the intriguing possibility that the clean images do not reside on the local peaks of the image priors learned in the diffusion models. This may make sense given the in memorization v.s. generalization trade-off~\citep{kadkhodaiegeneralization}. That is, a model simply encodes every clean image as the prior mode may not generalize well. 


\begin{table}[htbp]
\caption{$\ell_2$ distance measurements during DiffPure on CIFAR-10 ($\ell_\infty=8/255$).}
\label{tab:res-cifar10-l2}
\centering
\begin{tabular}{l|ccc}
\toprule[1.5pt]
{\textbf{Attack}} & \textbf{Init (\bm{$t=0$})} & \textbf{Max (\bm{$t=100$})} & \textbf{End (\bm{$t=200$})} \\
\midrule
{BPDA}       & 1.027 $\pm$ 0.023 & 8.976 $\pm$ 0.118 & 3.606 $\pm$ 0.607 \\
{BPDA-EOT}   & 1.072 $\pm$ 0.046 & 8.992 $\pm$ 0.116 & 3.607 $\pm$ 0.615 \\
{PGD (Full)} & 1.077 $\pm$ 0.040 & 8.980 $\pm$ 0.118 & 3.646 $\pm$ 0.614 \\
{PGD-EOT}    & 1.188 $\pm$ 0.072 & 8.999 $\pm$ 0.115 & 3.695 $\pm$ 0.617 \\
\midrule
{Uniform ($\epsilon=8/255$)}   & 1.004 $\pm$ 0.009 & 8.979 $\pm$ 0.113 & 3.598 $\pm$ 0.618 \\
{Uniform ($\epsilon=16/255$)}  & 4.015 $\pm$ 0.034 & 9.699 $\pm$ 0.124 & 3.823 $\pm$ 0.640 \\
{Uniform ($\epsilon=32/255$)}  & 8.030 $\pm$ 0.065 & 11.715 $\pm$ 0.145 & 5.258 $\pm$ 0.714 \\
{Uniform ($\epsilon=128/255$)} & 16.051 $\pm$ 0.129 & 17.622 $\pm$ 0.184 & 9.307 $\pm$ 0.581 \\
\bottomrule[1.5pt]
\end{tabular}
\end{table}

\begin{table}[htbp]
\caption{$\ell_\infty$ distance measurements during DiffPure on CIFAR-10 ($\ell_\infty=8/255$).}
\label{tab:res-cifar10-linf}
\centering
\begin{tabular}{l|ccc}
\toprule[1.5pt]
{\textbf{Attack}} & \textbf{Init (\bm{$t=0$})} & \textbf{Max (\bm{$t=100$})} & \textbf{End (\bm{$t=200$})} \\
\midrule
{BPDA}       & 0.031 $\pm$ 0.000 & 0.601 $\pm$ 0.051 & 0.313 $\pm$ 0.053 \\
{BPDA-EOT}   & 0.031 $\pm$ 0.000 & 0.603 $\pm$ 0.051 & 0.316 $\pm$ 0.053 \\
{PGD (Full)} & 0.031 $\pm$ 0.000 & 0.606 $\pm$ 0.050 & 0.317 $\pm$ 0.055 \\
{PGD-EOT}    & 0.031 $\pm$ 0.000 & 0.606 $\pm$ 0.050 & 0.319 $\pm$ 0.053 \\
\midrule
{Uniform ($\epsilon=8/255$)}   & 0.031 $\pm$ 0.000 & 0.605 $\pm$ 0.050 & 0.314 $\pm$ 0.052 \\
{Uniform ($\epsilon=16/255$)}  & 0.125 $\pm$ 0.000 & 0.647 $\pm$ 0.054 & 0.330 $\pm$ 0.055 \\
{Uniform ($\epsilon=32/255$)}  & 0.251 $\pm$ 0.000 & 0.735 $\pm$ 0.052 & 0.425 $\pm$ 0.059 \\
{Uniform ($\epsilon=128/255$)} & 0.502 $\pm$ 0.000 & 0.941 $\pm$ 0.057 & 0.623 $\pm$ 0.054 \\
\bottomrule[1.5pt]
\end{tabular}
\end{table}

Although the results above indicate that diffusion models are ineffective in removing small perturbations, it is possible that they may be more effective in removing noise induced by larger perturbations.
We performed the same $\ell_2$ distance analysis using three larger levels of uniform noises, ranging from $\epsilon=\{32,64,128\}/255$, to examine the model behavior under larger perturbations. As the noise level increases, the $\ell_2$ distances of the final purified states increase. Interestingly, the model transits from ``pushing-away'' to ``shrinkage'' under very large perturbations. 

\subsection{Distributional and semantic distances fail to explain robustness improvements}
\label{app:fid-clip}
\begin{figure*}[h]
\centering
\subfloat[FID distance after purification.]{
\includegraphics[width=0.45\columnwidth]{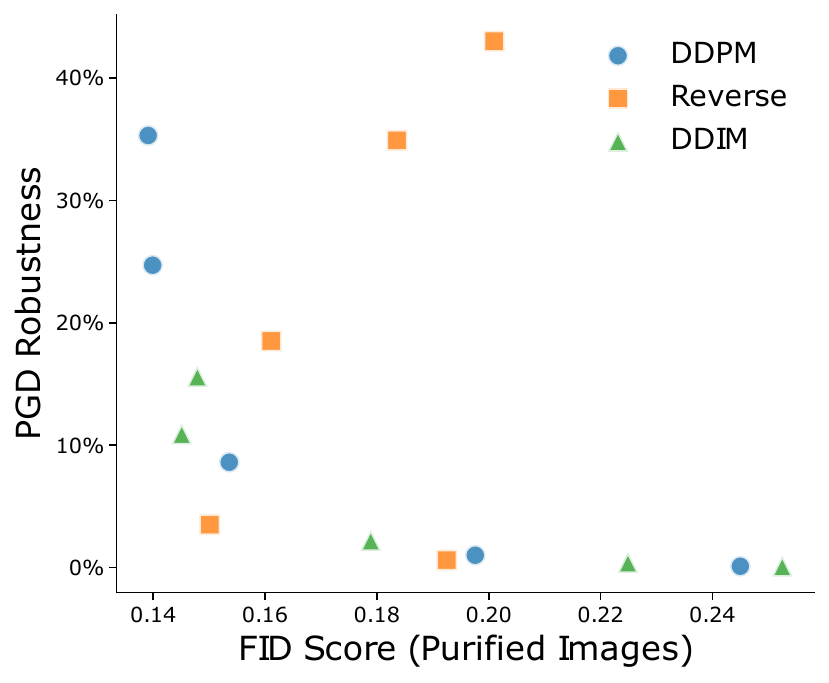}
\label{fig:fid}
}
\hspace{2em}
\subfloat[Cosine distance in the CLIP space.]{
\includegraphics[width=0.45\columnwidth]{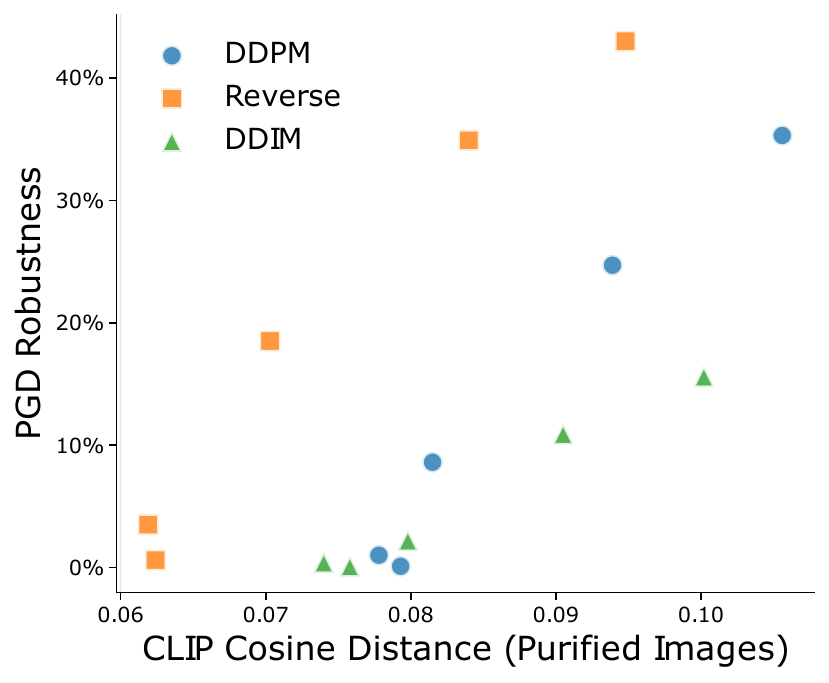}
\label{fig:clip}
}
\caption{\textbf{Distributional and semantic distances to the clean samples and adversarial robustness.} (a) Adversarial robustness and distributional distances (FID score) to clean sample distributions. (b) Adversarial robustness and semantic distances (cosine similarity in the CLIP space) to clean samples. For both metrics, we do not observe a monotonic relation with robustness, in contrast to our proposed compression rate, which exhibits a consistent monotonic relation (Fig.~5a).}
\label{fig:fid-clip}
\end{figure*}

\begin{table}[htbp]
\begin{minipage}[t]{0.49\linewidth}
\centering
\caption{FID to clean samples after purification.}
\label{tab:fid}
\small
\setlength{\tabcolsep}{4pt}
\begin{tabular}{l|ccccc}
\toprule[1.5pt]
{\textbf{Method}} & $t=10$ & $t=20$ & $t=50$ & $t=100$ & $t=150$ \\
\midrule
\multirow{1}{*}{DDPM}
& 0.245 & 0.198 & 0.154 & 0.140 & 0.139 \\
\multirow{1}{*}{Reverse}
& 0.192 & 0.150 & 0.161 & 0.184 & 0.201 \\
\multirow{1}{*}{DDIM}
& 0.253 & 0.225 & 0.179 & 0.145 & 0.148 \\
\bottomrule[1.5pt]
\end{tabular}
\end{minipage}%
\hfill
\begin{minipage}[t]{0.49\linewidth}
\centering
\caption{CLIP distance reduction after purification.}
\label{tab:clip-distance}
\small
\setlength{\tabcolsep}{4pt}
\begin{tabular}{l|ccccc}
\toprule[1.5pt]
{\textbf{Method}} & $t=10$ & $t=20$ & $t=50$ & $t=100$ & $t=150$ \\
\midrule
{DDPM}   & 0.0285 & 0.0249 & 0.0082 & -0.0162 & -0.0330 \\
{Reverse}  & 0.0456 & 0.0359 & 0.0100 & -0.0135 & -0.0272 \\
{DDIM} & 0.0327 & 0.0333 & 0.0195 & 0.0009  & -0.0131 \\
\bottomrule[1.5pt]
\end{tabular}
\end{minipage}
\end{table}


We further measured the cosine distance between the adversarial samples and their corresponding clean samples in the CLIP representation space, both before and after purification. As shown in Table~\ref{tab:clip-distance}, we observed a reduction in distance for smaller timesteps $t$, whereas for larger $t$ the distance increases.

Similarly, if we treat the purified CLIP distance to the clean samples as a robustness indicator, the results are shown in Fig.~\ref{fig:clip}.

Overall, neither FID nor CLIP semantic distances exhibit a consistent monotonic relationship with robustness, indicating that they are not reliable robustness indicators. This contrasts with the strong and consistent correlation observed with our proposed expected compression rate, further supporting our claim that robustness arises from a global compression of the image space rather than proximity to clean samples in FID or semantic space.

\subsection{Reliability analysis of the expected compression rates} We next analyze the reliability of the expected compression rates. Given the simplicity of the definition, the only hyperparameter involved is the magnitude of the random perturbation $\epsilon$. When $\epsilon$ is sufficiently small to make the first-order Taylor expansion valid, the exact magnitude of $\epsilon$ should not matter. Empirically, we used the standard magnitude of the adversarial attack for each dataset. To further examine the relationship between perturbation magnitude and CR, we measured the expected CR ranging from 1/4 to 4 times of the original adversarial magnitude on CIFAR-10. As shown in Table~\ref{tab:cr-epsilon-cifar10}, for small $\epsilon$ (up to 16/255), CR remains stable. Only when $\epsilon$ becomes noticeably large (e.g., 32/255), the CR begins to drift. This is expected as the Taylor approximation would break down for large perturbations. We also verified that the expected CR is stable across different random seeds (Table~\ref{tab:cr-seeds-cifar10}). Together, these results suggest that the expected CR is a simple yet remarkably stable quantity that reflects the intrinsic compression capability of diffusion models.

\begin{table}[t]
\begin{minipage}[t]{0.58\linewidth}
\centering
\caption{Compression rates with different perturbation scales.}
\label{tab:cr-epsilon-cifar10}
\small
\begin{tabular}{l|ccccc}
\toprule[1.2pt]
\textbf{Epsilon} & \textbf{2/255} & \textbf{4/255} & \textbf{8/255} & \textbf{16/255} & \textbf{32/255} \\
\midrule
Expected CR & 0.230 & 0.230 & 0.231 & 0.239 & 0.268 \\
Expected CR (std)    & 0.047 & 0.046 & 0.046 & 0.049 & 0.057 \\
\bottomrule[1.2pt]
\end{tabular}
\end{minipage}%
\hfill
\begin{minipage}[t]{0.40\linewidth}
\centering
\caption{Compression rates across random seeds.}
\label{tab:cr-seeds-cifar10}
\small
\begin{tabular}{l|ccc}
\toprule[1.2pt]
\textbf{Seed} & \textbf{0} & \textbf{123} & \textbf{295} \\
\midrule
Expected CR & 0.2306 & 0.2309 & 0.2312 \\
Expected CR (std)    & 0.0457 & 0.0481 & 0.0481 \\
\bottomrule[1.2pt]
\end{tabular}
\end{minipage}
\end{table}

%
%

\subsection{Adversarial attacks exploit nonlinear geometry}
\label{app:nonlinear}
While our proposed hyperspherical model suggested that the adversarial region can be well approximated by the behavior of a linear decision boundary and our empirical evaluation based on the sharpness of the choice transition in Fig.~3 supports this model, our evaluation also reveals that the slope of the transition is finite. This implies a slight deviation from a linear decision boundary at the scale of standard adversarial attack.  Conceivably, adversarial attacks likely exploit the such deviation from linearity to find the weakest directions. 
To further isolate the contribution of directional alignment from
other properties of adversarial perturbations, we construct
a controlled baseline using the same $\alpha$-interpolation
construction introduced above for the base classifier, now
with the diffusion system's test-time adversarial direction
$\boldsymbol{\eta}(\boldsymbol{\xi}_{\textrm{test}})$ as the
target: we linearly interpolate between
$\boldsymbol{\eta}(\boldsymbol{\xi}_{\textrm{test}})$ and a
random direction $\boldsymbol{\eta}_{\textrm{random}}$,
\begin{equation}
\hat{\boldsymbol{\eta}}(\alpha) = 
\frac{\alpha \cdot \boldsymbol{\eta}(\boldsymbol{\xi}_{
		\textrm{test}}) + (1-\alpha) \cdot 
	\boldsymbol{\eta}_{\textrm{random}}}
{\|\alpha \cdot \boldsymbol{\eta}(\boldsymbol{\xi}_{
		\textrm{test}}) + (1-\alpha) \cdot 
	\boldsymbol{\eta}_{\textrm{random}}\|_2} \cdot 
\|\boldsymbol{\eta}(\boldsymbol{\xi}_{\textrm{test}})\|_2,
\end{equation}
where $\ell_2$ normalization ensures that only the 
\emph{direction} varies across $\alpha$, not the perturbation 
magnitude. Note that the perturbations constructed here by the $\alpha$-interpolation match the $L_2$ norm (not the $L_{\infty}$ norm) of the adversarial attacks.
 Sweeping $\alpha \in [0, 1]$ traces out a smooth 
curve in (correlation, residual) space that captures what 
residual one would expect from a perturbation with a given 
correlation $\rho$. If that perturbation had no 
adversarial structure beyond its directional alignment with 
$\boldsymbol{\eta}(\boldsymbol{\xi}_{\textrm{test}})$. This 
curve therefore serves as a first-order, direction-only 
baseline.

\begin{figure}[t]
\centering
\includegraphics[width=0.5\linewidth]
{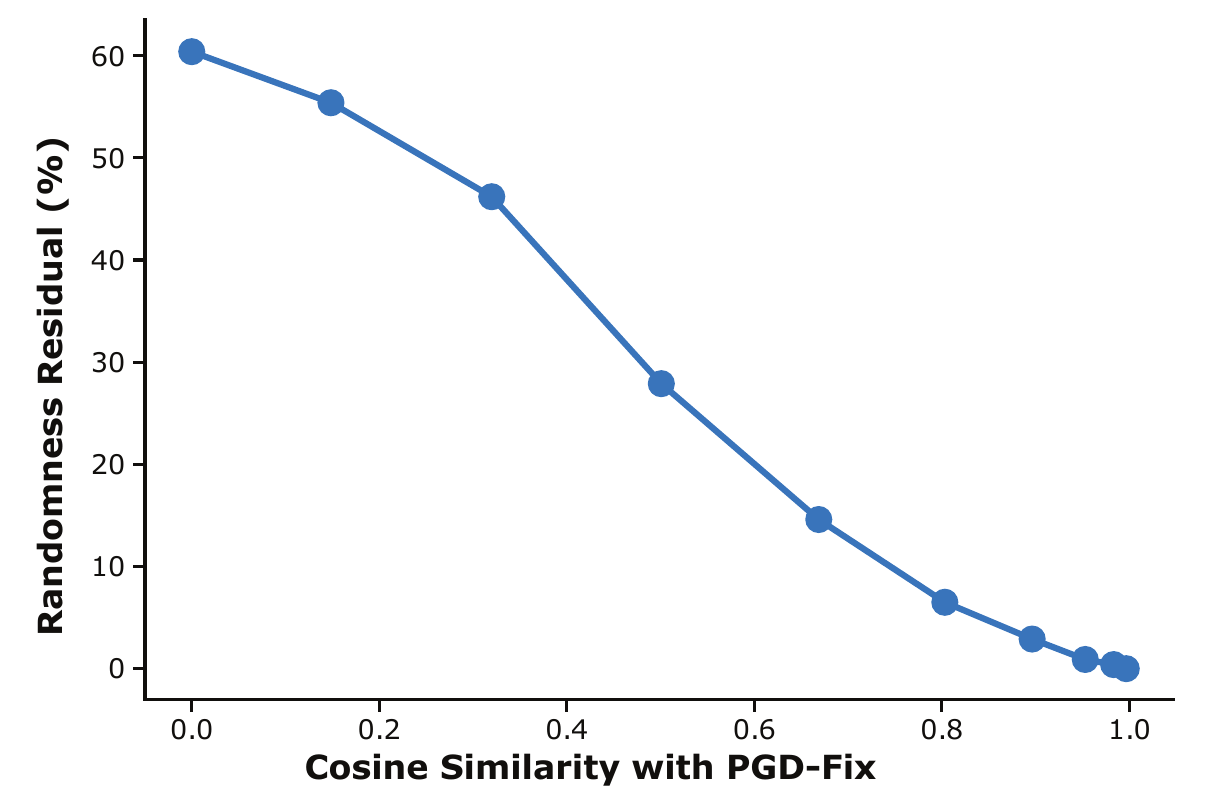}
\caption{
	\textbf{Robust residual as a function of perturbation 
		correlation with the fixed-randomness attack 
		(PGD-fix), based on the interpolation experiments.} The blue curve shows the residual obtained 
	by the $\alpha$-interpolation baseline 
	$\hat{\boldsymbol{\eta}}(\alpha) \propto 
	\alpha\cdot\boldsymbol{\eta}(\boldsymbol{\xi}_{\textrm{test}}) 
	+ (1-\alpha)\cdot\boldsymbol{\eta}_{\textrm{random}}$, 
	normalized to constant $\ell_2$ norm across all $\alpha$.
}
\label{fig:SI_robust-residual}
\end{figure}

The results are reported in (Fig.~\ref{fig:SI_robust-residual}). The robustness residual decreases smoothly with the cosine similarity between the actual attack and the attack calculated based on fixing randomness (PGD-fix). Furthermore, comparing these results to the data points in Fig.~4, it is evident that all non-fixed adversarial 
attacks---PGD with varying numbers of steps, PGD-EOT, and 
uniform noise---fall \emph{below} the interpolation curve in Fig.~\ref{fig:SI_robust-residual}. Since perturbation norm is held 
constant, this gap cannot be attributed to differences in the strength of the
attack. Instead, it reveals that adversarially 
optimized directions $\boldsymbol{\eta}(\boldsymbol{\xi}_{
\textrm{attack}})$ are more effective than direction-matched 
random baselines at the same correlation level: they exploit 
\emph{nonlinear structure} in the loss landscape that is 
invisible to first-order directional alignment alone. 

These results provide evidence suggesting that adversarial attacks exploit the deviation from the decision boundary. Despite this nonlinear property that is not captured by our hyperspherical cap model of the adversarial region, the randomness residual (relative to  the fixed-randomness condition) remains tightly related to the cosine similarity between the adversarial attack and the optimal attack (obtained based on fixing randomness).

\newpage
\section{Examples of adversarial and purified images with different compression rates}
\label{app:examples}
\begin{figure*}[h]
\centering
\subfloat[$t=100$.]{
\includegraphics[width=\columnwidth]{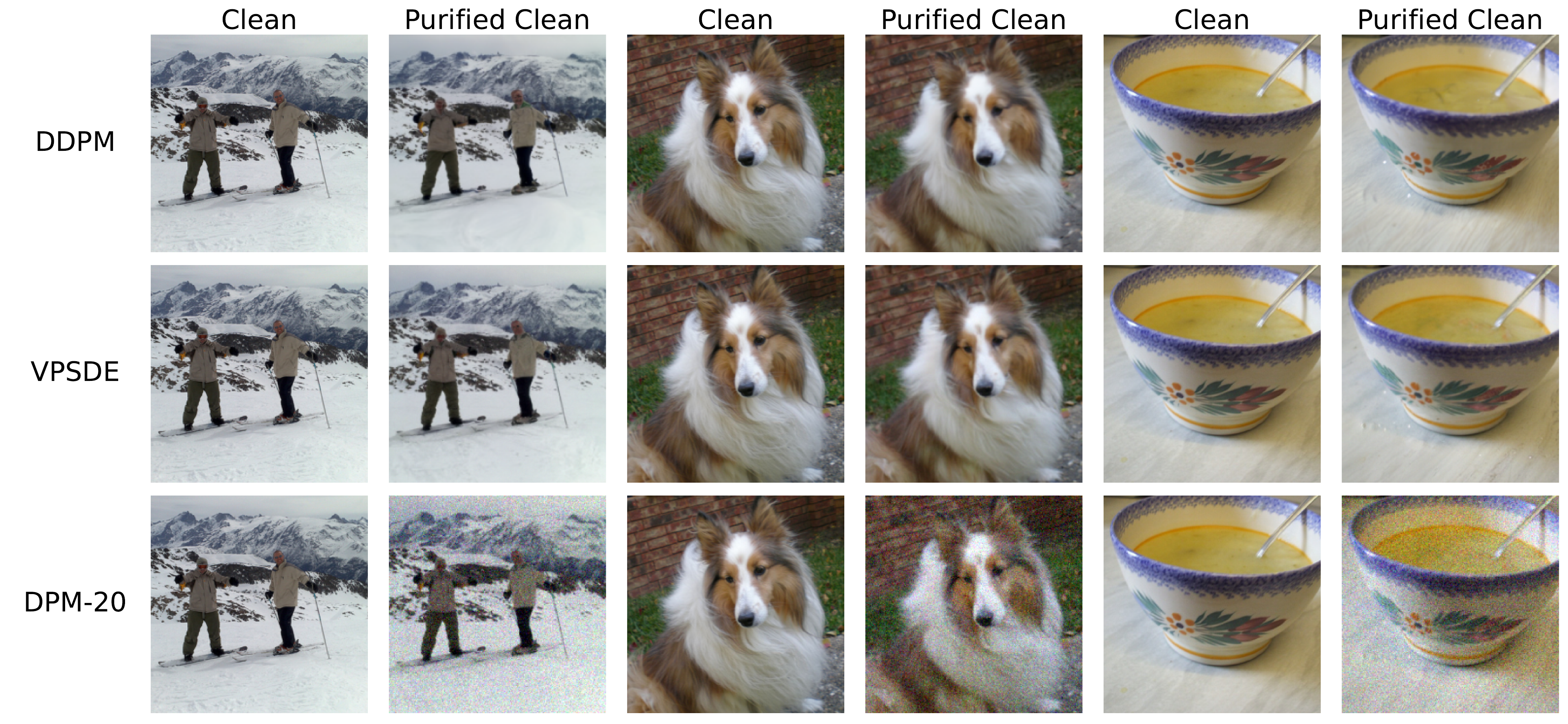}
}
\caption{Examples of adversarial and purified images on ImageNet.}
\label{fig:examples-imagenet}
\end{figure*}

\begin{figure*}[h]
\centering
\subfloat[DDPM.]{
\includegraphics[width=\columnwidth]{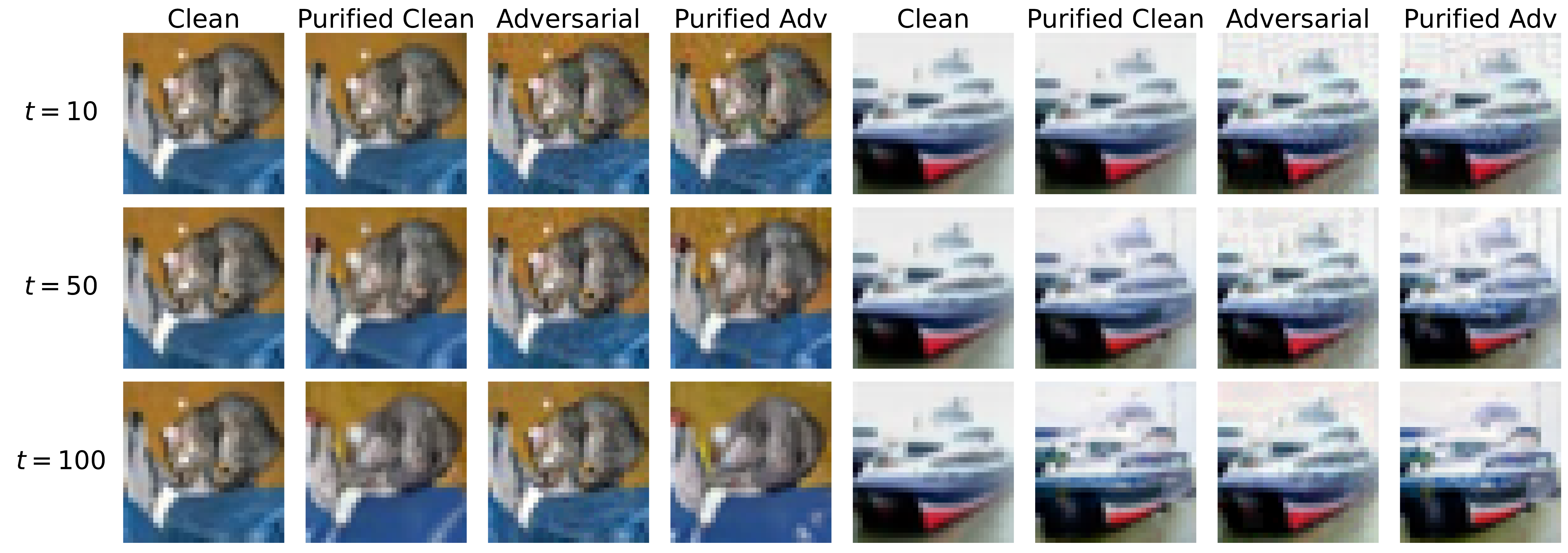}
}

\vspace{0.8em}

\subfloat[Reverse-only DDPM.]{
\includegraphics[width=\columnwidth]{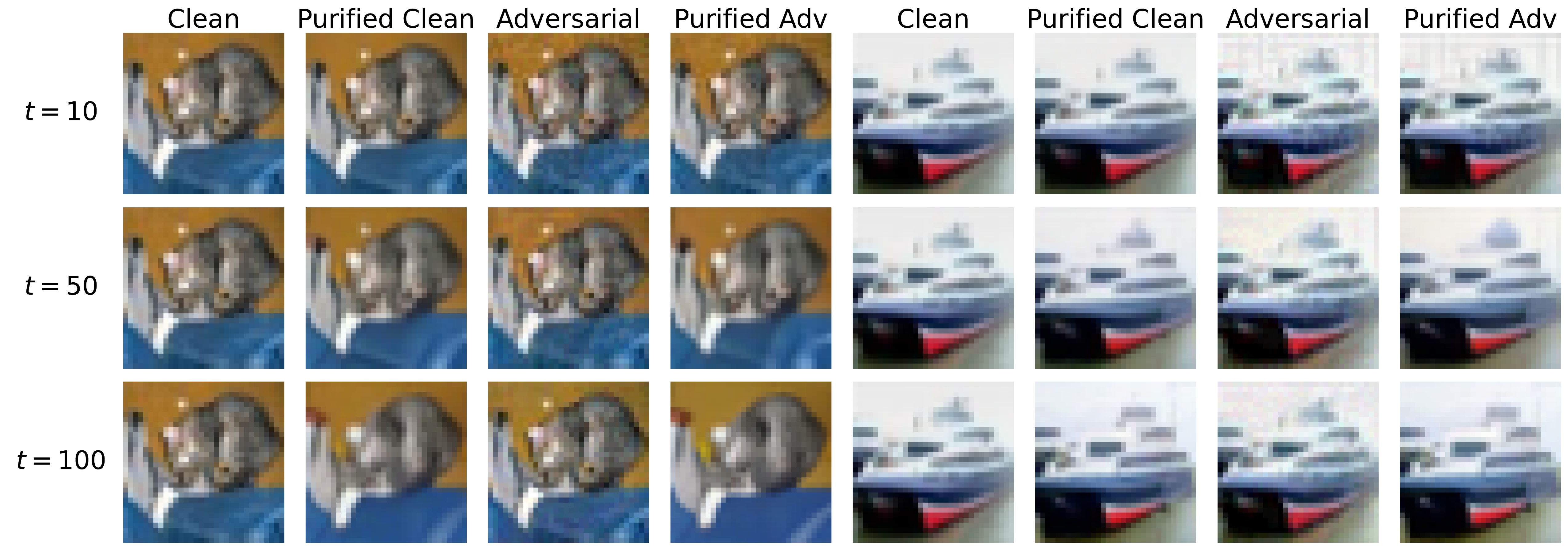}
}

\vspace{0.8em}

\subfloat[DDIM.]{
\includegraphics[width=\columnwidth]{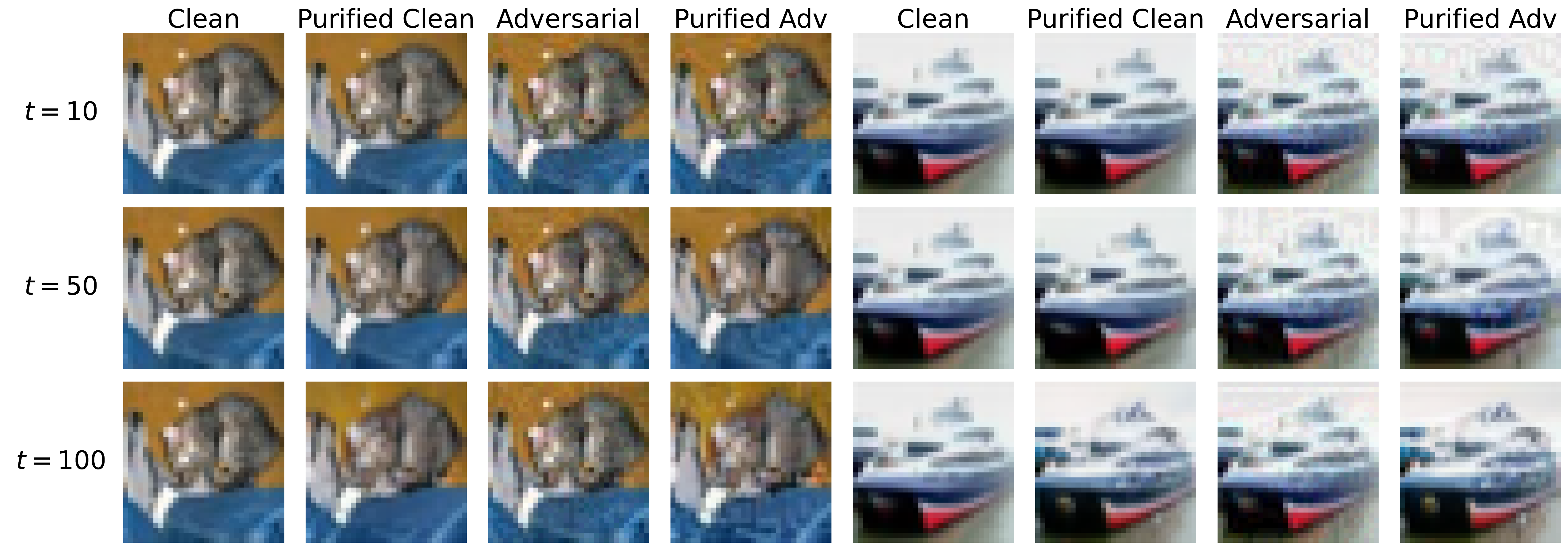}
}

\caption{Examples of adversarial and purified images on CIFAR-10.}
\label{fig:examples-cifar10}
\end{figure*}

\clearpage
\renewcommand\bibsection{\section*{Appendix References}}
\putbib[diffusion-robust]
\end{bibunit}

\end{document}